%% file: main.tex
\documentclass[10pt,twocolumn,letterpaper]{article}
\pdfoutput=1

\usepackage[dvipsnames]{xcolor}

\usepackage[pagenumbers]{cvpr} 

\definecolor{cvprblue}{rgb}{0.21,0.49,0.74}
\usepackage[pagebackref,breaklinks,colorlinks,citecolor=cvprblue]{hyperref}

\def\paperID{4563} 
\def\confName{CVPR}
\def\confYear{2024}

\title{WordRobe: Text-Guided Generation of Textured 3D Garments}

\author{Astitva Srivastava$^{1}$\\
\and
Pranav Manu$^{1}$
\and
Amit Raj$^{2}$
\and
Varun Jampani$^{3}$
\and
Avinash Sharma$^{1,4}$
\and
$^1$IIIT Hyderabad, India\\
\and
$^2$Google Research\\
\and
$^3$Stability AI\\
\and
$^4$IIT Jodhpur, India\\
}
\begin{document}
\twocolumn[{ %
\renewcommand\twocolumn[1][]{#1}%
\maketitle
\begin{center}
    \centering
    \captionsetup{type=figure}
    \includegraphics[width=\linewidth]{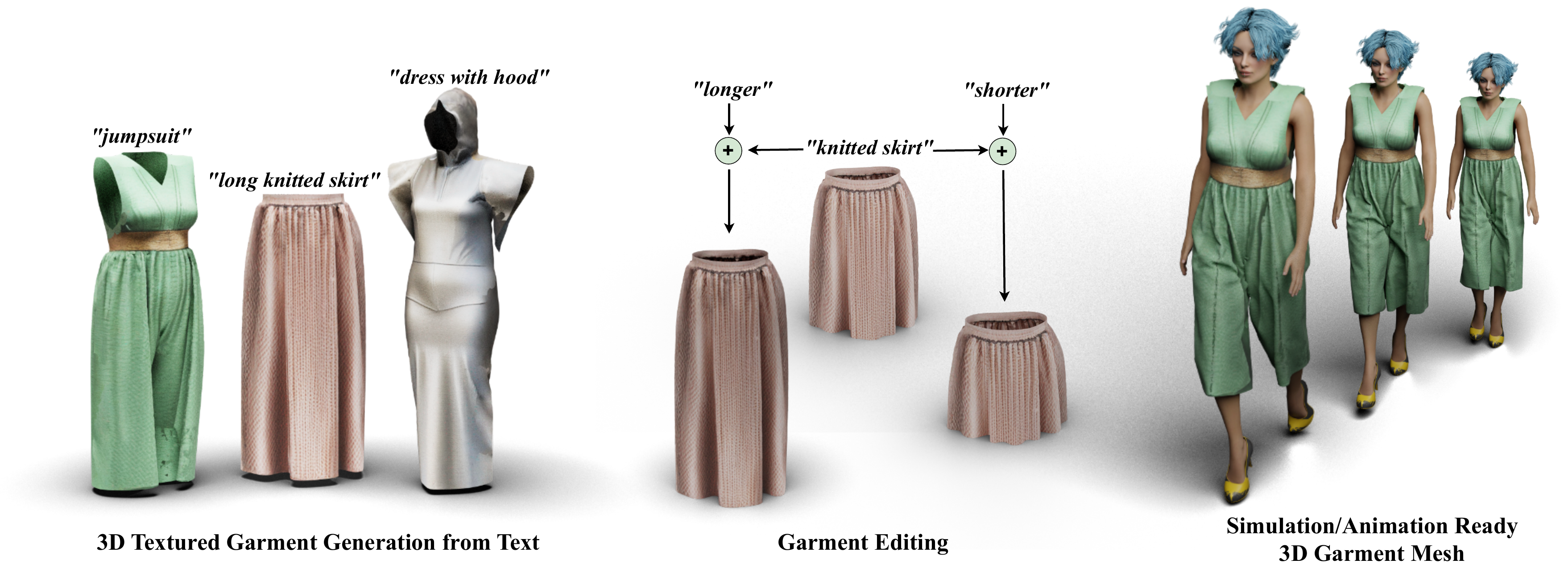}
    \captionof{figure}{Text-guided generation and editing of 3D textured garments using \textbf{WordRobe}.}
    \label{fig:teaser} 
\end{center}
}]
\input{sec/0_abstract}    
\input{sec/1_intro}
\input{sec/2_related_work}
\input{sec/4_method}
\input{sec/5_evaluation}
\input{sec/6_conclusion}
{
    \small
    \bibliographystyle{unsrtnat}
    \bibliography{main}
}

\end{document}

%% file: sec/0_abstract.tex
\begin{abstract}
In this paper, we tackle a new and challenging problem of text-driven generation of 3D garments with high-quality textures. We propose, \textbf{\textit{WordRobe}}, a novel framework for the generation of \textit{unposed} \& textured 3D garment meshes from \textit{user-friendly} text prompts. 
We achieve this by first learning a latent representation of 3D garments using a novel coarse-to-fine training strategy and a loss for latent disentanglement, promoting better latent interpolation. Subsequently, we align the garment latent space to the CLIP embedding space in a weakly supervised manner, enabling text-driven 3D garment generation and editing. For appearance modeling, we leverage the zero-shot generation capability of ControlNet
to synthesize view-consistent texture maps in a single feed-forward inference step, thereby drastically decreasing the generation time as compared to existing methods. We demonstrate superior performance over current SOTAs for learning 3D garment latent space, garment interpolation, and text-driven texture synthesis, supported by quantitative evaluation and qualitative user study. The unposed 3D garment meshes generated using \textit{WordRobe} can be directly fed to standard cloth simulation \& animation pipelines without any post-processing.
\href{https://wordrobe24.github.io/WordRobe_Page/}{\textcolor{cyan}{[Link to project page]}}
\end{abstract}

%% file: sec/1_intro.tex
\section{Introduction}
\label{sec:intro}
%
Recent advances in 3D content creation from textual description has large implications for modelling the virtual world. This includes a wide variety of assets such as objects\cite{raj2023dreambooth3d}, scenes\cite{zhang2023text2nerf}, as well as humans\cite{kolotouros2023dreamhuman}. Automated content creation has also fueled interest in generating 3D garments for applications in 3D virtual try-on, clothing for human avatars, gaming \& animated characters, and AR/VR experiences. The 3D garments are typically represented as textured meshes to model the underlying surface geometry \& appearance. 
However, creating large-scale 3D garments is prohibitively expensive, primarily due to the huge diversity in the shape, style, and appearance of the garments.
Noise-free, \textit{unposed} (i.e. in canonical/T-pose) 3D garment modelling is important for direct integration into simulation and animation pipelines. To achieve this, traditional approaches either employ design tools for manual garment creation (e.g., CLO\cite{clo3d}) or capture digital replicas of real garments via high-end scanners (e.g., Artec\cite{artec}). However, such approaches require significant design effort, and are expensive and difficult to scale up. Thus, there is an acute need to develop a scalable learning-based solution for automated 3D garment creation that effectively models shape, style \& appearance of various garments.
%
%
%
%
%

A variety of deep learning-based methods aim to digitize/reconstruct 3D garments from images, which can be broadly divided into two categories based on the garment representation, namely parametric or non-parametric. Parametric garment reconstruction methods \cite{jiang2020bcnet, corona2021smplicit} are restricted to tight-fitting and limited clothing designs due to reliance on garment templates derived from an underlying parametric human model (e.g. SMPL~\cite{SMPL:2015}). Nevertheless, their parametric nature supports high-quality texture maps and editing of the pose, size \& shape \cite{deepcloth_su2022}. On the other hand, non-parametric learning-based methods \cite{zhu2022registering, Srivastava2022xClothET} can model garments of various styles and appearances (within training distribution). However, they yield \textit{posed} geometry \& low-quality textures, making the output garment ill-suited for direct integration into standard graphics pipelines. Furthermore, these methods offer no control over the shape and pose editing of the 3D garment.
%
%
An alternate approach for non-parametric 3D garment modeling is effectively demonstrated in DrapeNet\cite{de2023drapenet}, highlighting the capability of MLPs to learn the shape distribution of 3D garments by encoding them in a latent space. This enables shape editing via latent interpolation. However, there is no support for texture which is crucial for modelling high-quality appearance details. Additionally, the generation from the garment latent space is uncontrolled and the latent manipulation is also defined by explicit per-component labels, which require significant annotation effort in case of a large variety of garments made of different components. Therefore, a controllable way to generate and edit 3D garments via intuitive inputs (e.g. images or text prompts) is desirable.

Recent text-to-3D methods \cite{raj2023dreambooth3d, wang2023prolificdreamer, Chen_2023_ICCV, sun2023dreamcraft3d} allow generation of generic 3D objects via \textit{\textit{user-friendly}} text prompts, eschewing the need for 3D  modelling and artistic expertise. However, when employed for generating 3D garments, the surface quality of the generated garment mesh is subpar as compared to the methods trained specifically to model garments. While the overall quality of text-to-3D methods is expected to improve as research progresses, the inherent 3D representations used by these methods have certain limitations in terms of representing 3D garments with open (non-watertight) surfaces and also lack support for editing or manipulation. Additionally, the majority of these methods rely on a multiview optimization process, which is computationally expensive and slow. 

To this end, we propose \textbf{\textit{WordRobe}}, a text-driven textured 3D garment generation framework. As shown in \autoref{fig:teaser}, \textit{WordRobe} generates high-quality \textit{unposed} 3D garment meshes with photorealistic textures from \textit{\textit{user-friendly}} text prompts.
%
%
We achieve this by first learning a latent space of 3D garments using a novel two-stage encoder-decoder framework in a coarse-to-fine manner, representing the 3D garments as unsigned distance fields (UDFs). We also introduce an additional loss function to further disentangle the latent space, promoting better interpolation. We devise a new metric to quantitatively study the effect of the proposed loss function on garment interpolation. Once the garment latent space is learned, we train a mapping network to predict garment latent codes from CLIP embeddings. This allows CLIP-guided exploration of the latent space, enabling text-driven 3D garment generation and editing. For training the aforementioned mapping network, we develop a novel \emph{weakly-supervised} training scheme that eliminates the need for explicit manual text annotations. For text-guided texture synthesis, we leverage the capabilities of pretrained T2I models for generating diverse textures. Unlike existing multiview optimization-based methods \cite{yu2023texture, richardson2023texture, chen2023text2tex}, which are slow and expensive, we render the front \& back depth maps of the 3D garments side-by-side in a single image and pass this image to ControlNet\cite{zhang2023adding} for a depth-conditioned image generation. This novel approach enables text-driven texture synthesis in a single feed-forward step, saving time while outperforming existing SOTA\cite{chen2023text2tex} in maintaining view consistency.
To the best of our knowledge, our method is the first one to enable the text-driven generation of high-fidelity 3D garments with diverse textures.
In summary, our major contributions are as follows:
\begin{itemize}
    \item A novel framework and training strategy for text-driven 3D garment generation via a garment latent space.
    \item A new disentanglement loss for promoting better separation of concepts in the latent space and a new metric to assess its performance.
    \item An optimization-free (single feed-forward) text-guided texture synthesis method that is both superior and efficient as compared to existing SOTA.
\end{itemize}
\noindent
We also extend the 3D garments dataset proposed in \cite{KorostelevaGarmentData} with diverse high-quality textures and corresponding text prompts, using the proposed approach. We plan to publicly release the dataset and code to further accelerate research in this space.

%% file: sec/2_related_work.tex
\section{Related Work}
\begin{figure*}[t!]
    \centering
    \includegraphics[width=\textwidth]{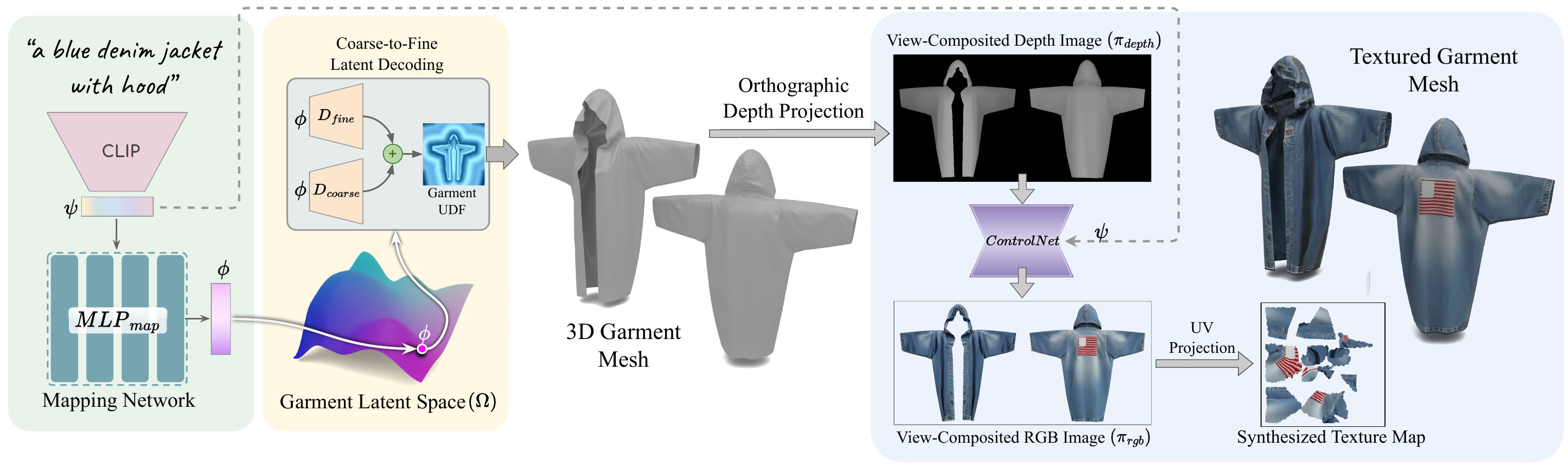}
    \caption{Overview of the proposed method for text-guided 3D garment generation.} 
    \label{fig:pipeline}
\end{figure*}
\noindent
\textbf{3D Garment Digitization:} 
Researchers have proposed several deep-learning methods\cite{bhatnagar2019mgn, jiang2020bcnet, Zhu2020DeepFA, zhu2022registering, Srivastava2022xClothET} that attempt to digitize /reconstruct 3D garments from images. 
Methods such as \cite{saito2019pifu, saito2020pifuhd, zheng2021pamir, xiu2023econ} use neural implicit representations (e.g. occupancy field, SDF, etc.) to model 3D clothed humans from monocular or sparse multi-view images, in a supervised learning setup. But, they fail to model garments separately from the body. ReEF\cite{zhu2022registering} learns explicit boundary curves and segmentation field to model garments separately, while xCloth\cite{Srivastava2022xClothET} proposed to use an alternate and efficient representation to achieve the same, while also obtaining texture maps. However, all these methods rely on high-quality real-world clothed human datasets\cite{3DHumansDataset} which have a limited diversity in terms of style \& appearance since capturing such datasets at a large scale is expensive. Moreover, the reconstructed garments are \textit{posed} according to the underlying body and have a sub-optimal surface quality.
Another line of works takes inspiration from real-world garment creation and proposes both analytical\cite{10.1145/3528223.3530145} \& neural methods\cite{KorostelevaGarmentData} for procedurally generating production-ready \textit{unposed} 3D garments. However, such approaches rely on sophisticated sewing patterns which are not intuitive to design. 
Some of the recent approaches\cite{corona2021smplicit, deepcloth_su2022} avoid panel-based generation by building upon parametric human body templates (e.g., SMPL\cite{SMPL:2015}) to generate parametric garments, however, they usually model tight-fitted garments with limited texture support. \\
\\
\noindent
\textbf{Text-to-3D Generation:}
Recently, various text-to-3D methods have been proposed \cite{raj2023dreambooth3d, wang2023prolificdreamer, Chen_2023_ICCV, sun2023dreamcraft3d, poole2022dreamfusion,gao2023textdeformer, shen2021deep} which leverage powerful imaginative capabilities of text-to-image (T2I) diffusion models \cite{rombach2022highresolution}, combined with popular 3D representations (NeRF\cite{mildenhall2020nerf}, DMTet\cite{shen2021deep}, etc.) to generate 3D objects from the text prompts. However, most of these methods rely on a multiview optimization process which is computationally expensive. Moreover, NeRF-like representations are not suitable for modelling complex open garment surfaces, hence the output geometry quality is not sufficient to be directly integrated into a standard graphics pipeline. Additionally, these methods lack support for controllable manipulation or editing of the generated 3D mesh.
\\
\\
\noindent
\textbf{Text-Guided Texture Generation:}
Recently, Text-to-Image(T2I) Diffusion Models\cite{ho2020denoising} have garnered significant interest which has led to works like \cite{metzer2022latent, khalid2022clipmesh, chen2023text2tex, kant2023invs} for synthesizing 2D UV texture map for a given 3D mesh. The majority of these methods optimize the CLIP objective between the input text prompt and multiview images generated from a pretrained denoising diffusion model\cite{rombach2022highresolution}. However, the output resolution is low, resulting in poor texture quality.
Current SOTA method Text2Tex\cite{chen2023text2tex} utilizes a depth-aware image inpainting diffusion model to progressively fill in a high-resolution texture on a mesh conditioned on a text prompt. 
However, the progressive nature of texture filling makes this method relatively time-consuming. The resultant texture map also suffers from view inconsistencies because the denoising process across different views generates different images.
On the contrary, we propose a texture generation method that generates view-consistent textures by generating all the views at once in a single feed-forward step. Note that our approach differs significantly from \cite{kant2023invs, zhao2023efficientdreamer, shi2023mvdream} since these methods explicitly train the Stable Diffusion\cite{rombach2022highresolution} with orthogonal views on 3D datasets with limited textural diversity (e.g. Objaverse\cite{deitke2022objaverse}), whereas we leverage the zero-shot generation capabilities and the newly identified property of the pretrained ControlNet\cite{zhang2023adding}, enabling arbitrarily diverse textures.
%
%
%
%

%% file: sec/4_method.tex
\section{Method}
\begin{figure*}[t!]
    \centering
    \includegraphics[width=\textwidth]{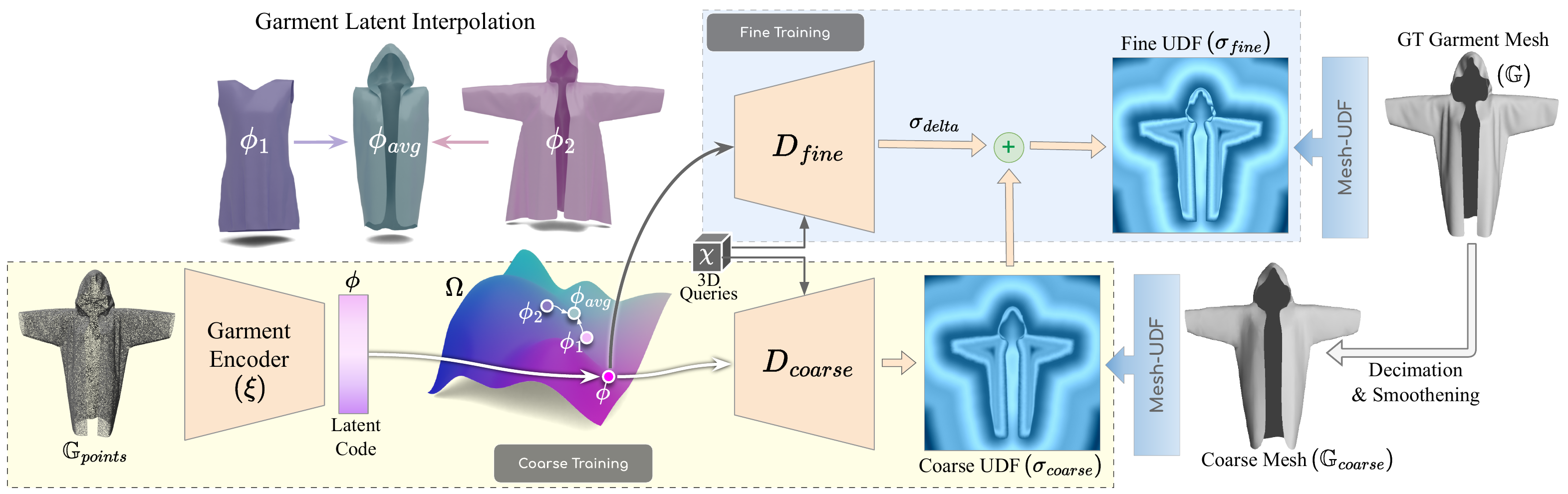}
    \caption{The proposed coarse-to-fine training strategy for learning garment latent space.}
    \label{fig:gargen}
\end{figure*}
%
We propose \textit{WordRobe}, a method to generate different types of 3D garments with openings (armholes, necklines etc.) and diverse textures via \textit{\textit{user-friendly}} text prompts. To achieve this, we incorporate three novel components in \textit{WordRobe} $-$ \textbf{3D garment latent space} (${\Omega}$) which encodes \textit{unposed} 3D garments as latent codes, (Sec.\ref{sec:latent_space}); \textbf{Mapping Network} ($MLP_{map}$) which predicts garment latent code from input text prompt  (Sec.\ref{sec:mlp_mapping}); and \textbf{Text-guided texture synthesis} to generate high-quality diverse texture maps for the 3D garments (Sec.\ref{sec:texture_syn}). We provide an overview of the proposed method in \autoref{fig:pipeline}. At inference time, given an input text prompt, we first obtain its CLIP embedding $\psi$, which is subsequently passed to $MLP_{map}$ to obtain the latent code $\mathbf{\phi}$ $\in$ $\Omega$. We further perform two-step latent decoding of $\mathbf{\phi}$ to generate the 3D garment as UDF, and extract the UV parametrized mesh representation for the same. Finally, we perform text-guided texture synthesis in a single feed-forward step by leveraging ControlNet\cite{zhang2023adding} to obtain the textured 3D garment mesh.

\begin{figure*}[t!]
    \centering
    \includegraphics[width=\linewidth]{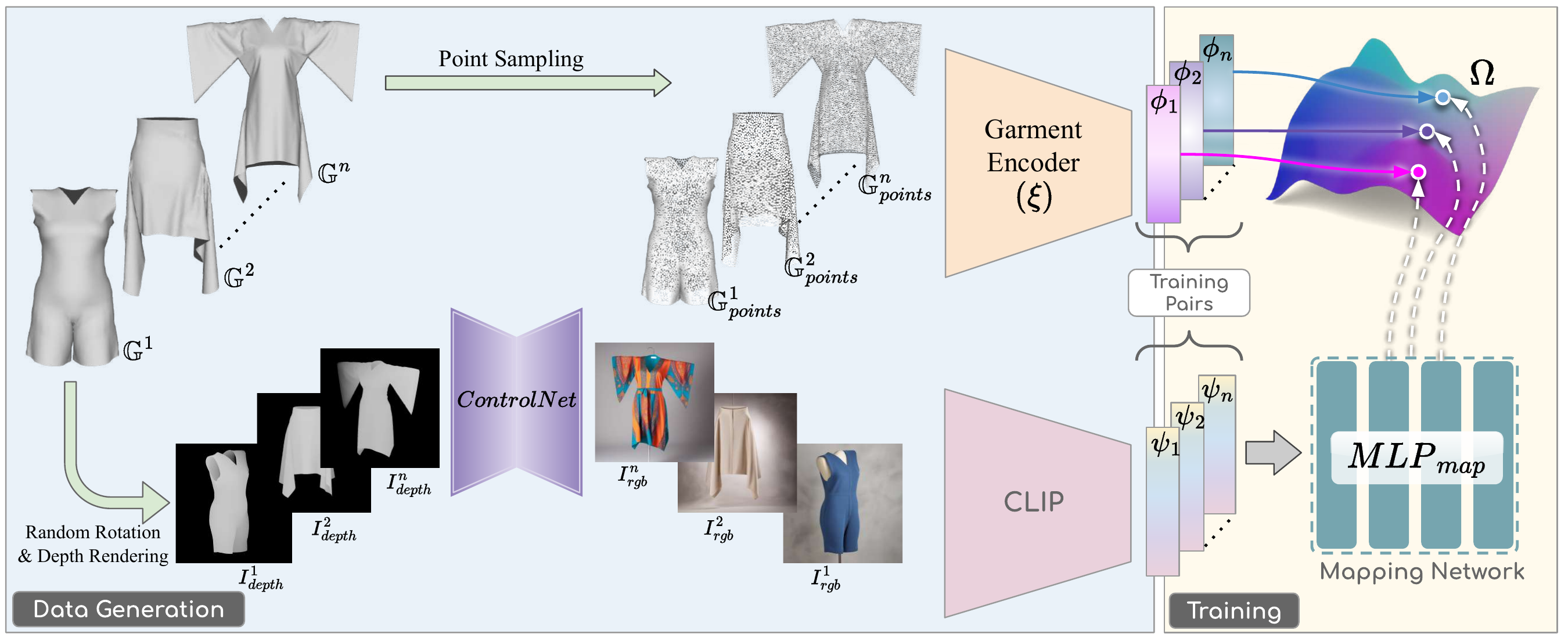}
    \caption{Automated training data generation \& weakly supervised training of $MLP_{map}$.}
    \label{fig:datagen}
\end{figure*}
\subsection{3D Garment Latent Space}
\label{sec:latent_space}
We propose to learn a latent space for the \textit{unposed} 3D garments using a novel two-stage encoder-decoder framework. Inspired by DrapeNet\cite{de2023drapenet}, we adopt the Unsigned Distance Function (UDF) to represent the open garment surfaces. We employ DGCNN\cite{dgcnn} as the garment encoder ($\xi$) to embed a variety of 3D garments into a latent representation by aggregating the multi-scale point features into a unified global embedding. As shown in \autoref{fig:gargen}, given a 3D garment mesh $\mathbb{G}$, we sample the points on the surface of the mesh and pass them to the encoder $\xi$. The output of the encoder is a $k=32$ dimensional garment latent code $\mathbf{\phi} \in \Omega$, where $\Omega$ is the garment latent space, i.e. 
%
   $ \mathbf{\phi}$ $=$ $\xi(\mathbb{G}_{points})$.
%
We use coordinate-based MLP\cite{mescheder2019occupancy} as the decoder, which takes $\mathbf{\phi}$ as input and decodes it into the UDF representation of the corresponding garment. However, we observe that a single decoder is not suited for learning both a regularized latent space and at the same time, modelling high-frequency details, such as wrinkles and pleats (see \autoref{fig:qual_ablate}, Sec.\ref{sec:qual_ablate}). Therefore, we propose to use two MLP decoders, $D_{coarse}$ and $D_{fine}$ to focus on two distinct objectives. Given a set of \textbf{$m$} query points, $\chi \in \mathbb{R}^{m\times3}$, defined over a 3D grid, $D_{coarse}$ predicts smooth (coarse) unsigned distance value for every query point according to the underlying geometry, conditioned on latent code $\mathbf{\phi}$. While $D_{fine}$ predicts residual change in the output of $D_{coarse}$ to capture finer details, i.e., 
%
%
%
\begin{equation}
    \sigma_{fine} = D_{coarse}(\mathbf{\phi}) + D_{fine}(\mathbf{\phi}) = \sigma_{coarse} + \sigma_{delta}
\label{eq:sigma-fine}
\end{equation}
%
%
We use Mesh-UDF\cite{guillard2022udf} to convert the 3D garment meshes into UDFs, which acts as ground truth for training $D_{fine}$. Similarly, for training $D_{coarse}$ we first decimate the 3D meshes, apply Laplacian Smoothing and pass it to Mesh-UDF to obtain ground truth coarse UDFs. 

As shown in \autoref{fig:gargen}, we train the proposed framework in two stages, where we first jointly train encoder $\xi$ and decoder $D_{coarse}$ to learn a rich garment latent space while decoding the latent codes into coarse UDF representations. We adopt distance loss ($\mathcal{L}_{dist}$) and gradient loss ($\mathcal{L}_{grad}$) from \cite{de2023drapenet}, where $\mathcal{L}_{dist}$ is formulated as BCE loss between the predicted and ground truth UDF values (normalized and clipped in the range $[0,1]$) and $\mathcal{L}_{grad}$ is the $L2$ distance between the gradients of predicted and ground truth UDFs. 
During training, we minimize $\mathcal{L}_{dist}$ and $\mathcal{L}_{grad}$ for each 3D query point $\mathbf{x} \in \chi$.
In order to have a more organized and disentangled latent space, we introduce a disentanglement loss $\mathcal{L}_{latent}$, which encourages the batch covariance $\mathbf{\Sigma}_b$  of the individual dimensions of latent vectors to be an identity matrix and is defined for a batch $b$ as follows:
\begin{equation}
    \mathcal{L}_{latent} = \mathbf{\Sigma}_{b} - \mathbf{I}_{k}
    \label{eq:batch_latent_loss}
\end{equation}

\begin{equation}
\mathbf{\Sigma}_b = 
\begin{bmatrix}
var(\mathbf{l}_1) & covar(\mathbf{l}_1, \mathbf{l}_2) & \cdots & covar(\mathbf{l}_1,\mathbf{l}_k) \\
covar(\mathbf{l}_2,\mathbf{l}_1) & var(\mathbf{l}_2) & \cdots & covar(\mathbf{l}_2,\mathbf{l}_k) \\
\vdots & \vdots & \ddots & \vdots \\
covar(\mathbf{l}_k,\mathbf{l}_1) & covar(\mathbf{l}_k, \mathbf{l}_2) & \cdots & var(\mathbf{l}_k) \\
\end{bmatrix}
\end{equation}
\noindent
where, $\mathbf{l}_i$ = \{$\mathbf{\phi}^1_{i}, \mathbf{\phi}^2_{i},...,\mathbf{\phi}^q_{i}; 1 \leq i \leq k$\} ($q$ is the batch size), $\phi_i$ is the $i^{th}$ dimension of the latent vector $\phi$, and $\mathbf{I}_k$ in \autoref{eq:batch_latent_loss} is $k \times k$ identity matrix.

In other words, $\mathcal{L}_{latent}$ enforces dimensions of latent vector $\phi$ to be as independent of each other as possible, allowing $\xi$ to encode the most prominent shape characteristics of the garments across different categories in the latent space $\Omega$. This results in a more organized latent space, where manipulation of the latent vector along a single (or very few) dimension(s) might be sufficient to have a desirable shape change in the 3D garment.
The respective loss functions for coarse and fine training stages are:
\begin{equation}
\begin{aligned}
\label{eq:coarse}
    \mathcal{L}_{coarse} = \lambda_{dist}\mathcal{L}_{dist} + \lambda_{grad}\mathcal{L}_{grad} + \lambda_{latent}\mathcal{L}_{latent} \\
    \mathcal{L}_{fine} = \lambda_{dist}\mathcal{L}_{dist} + \lambda_{grad}\mathcal{L}_{grad} \hspace{6.9em}
\end{aligned}
\end{equation}
\noindent
Minimizing the above losses results in a latent space $\Omega$ where we can randomly sample a latent vector $\phi$ and perform a two-step decoding to generate a 3D garment UDF associated with $\phi$. 3D garment mesh is extracted by running a modified version of Marching Cubes proposed in \cite{de2023drapenet}.
\subsection{CLIP-Guided 3D Garment Generation}
\label{sec:mlp_mapping}
We propose a novel weakly-supervised training scheme to align CLIP's latent space to the garment latent space $\Omega$. Given a text prompt, we first pass it through CLIP's text encoder to get an embedding $\psi$. We employ a mapping network $MLP_{map}$ that takes $\psi$ as input and predicts a garment latent code $\phi$.

In order to train $MLP_{map}$, a large set of garment latent codes and corresponding text prompts (to get corresponding CLIP embeddings) are required. To avoid explicit text annotations, we propose an automated way of generating training pairs (latent codes and CLIP embeddings).
As illustrated in \autoref{fig:datagen}, given a set of 3D garments $\mathbb{G}_{train}=\{\mathbb{G}^i | 1 \leq i \leq N\}$, we randomly rotate each garment mesh $\mathbb{G}^i$, render a depth map $I_{depth}^i$, and pass it to a depth-conditioned ControlNet\cite{zhang2023adding} to generate a garment image $I_{rgb}^i$, guided by a garment agnostic \textit{template prompt} $-$\textbf{\textit{``a garment made of }\{MATERIAL\}\textit{, with }\{COLOR\}\textit{ colors''}}.
We pick predefined values for {MATERIAL}$=$\{silk, cotton, wool, leather\} \& {COLOR}$=$\{vibrant, dull, bright, shiny, matte\} at random to construct the \textit{template prompt}. We add additional prompts (e.g. high-quality, realistic, photoreal, etc.) to ensure that ControlNet produces high-quality images which are then manually verified. The generated image $I_{rgb}^i$ is then passed to the CLIP's image encoder to generate clip embedding $\psi_{i}$. Concurrently, we sample points on the surface of every garment mesh $\mathbb{G}^i$ and feed them to the garment encoder $\xi$ to get corresponding latent code $\phi_{i}$. This technique eliminates the need for explicit manual text annotations for training the mapping network which is a huge benefit due to the lack of any such dataset.

Once all the corresponding pairs of $\psi_{i}$ and $\phi_{i}$ are generated, we train $MLP_{map}$ by minimizing the $L1$ loss between the $MLP_{map}$'s prediction and corresponding $\phi_{i}$. During inference, the mapping network $MLP_{map}$ takes the CLIP embedding $\psi$ of a text prompt and predicts a latent vector $\phi$, on which two-step decoding is performed to generate the 3D garment (as shown in \autoref{fig:pipeline}). This novel strategy enables taming the garment's latent space via text prompts.
%
\subsection{Texture Synthesis}
\label{sec:texture_syn}
Once we have the extracted 3D garment mesh, our next aim is to generate high-quality appearance and store it in the form of a UV texture map guided by the same input text prompt. Though UV parametrization is suitable for storage and fast rendering of the mesh, it is not trivial to generate textures directly in the UV space, as UV parametrization introduces seams, disturbing the spatial arrangement of the mesh primitives, which are organized differently in UV space for different meshes and do not carry any semantic meaning to help learning. Thus, we propose a novel strategy to synthesize textures from text prompts. 

Existing SOTA methods\cite{khalid2022clipmesh, kant2023invs} adopt Text-to-Image diffusion models in a multiview optimization framework, with the aim of having similar generations (in terms of colors, lighting, etc.) across different views while minimizing the CLIP objective. However, this approach is time-consuming and does not always guarantee view consistency as shown in \autoref{fig:ours_qual_tex2text}, as a small change in control (here, viewpoint) can drastically change the generated image. We, on the other hand, identify a highly useful property of ControlNet\cite{zhang2023adding}, which allows us to maintain consistency across different viewpoints of 3D garments in a single generation. More specifically, we empirically observed that if we composite multiview depth maps of a 3D object in a single image and pass it to ControlNet, the generated RGB image (guided by an input text prompt) tends to have consistent color values and lighting information across different views of that object in the image.

Leveraging the aforementioned property of ControlNet, we devise our text-driven texture synthesis methodology as follows. We first perform depth rendering of 3D garment mesh in two views $-$ front and back, and combine them together to obtain a view-composited depth image $\pi_{depth}$ as shown in \autoref{fig:pipeline}. Though any number of views can be used, We use front-back as a natural choice for partitioning a 3D garment to reduce the number of visible seams on the mesh, and also to maintain resolution. We use orthographic projection for rendering, as perspective projection leads to more information loss across tangential regions. $\pi_{depth}$ is then passed to ControlNet which generates a $1024\times1024$ view-composited RGB image $\pi_{rgb}$, conditioned on CLIP embedding $\psi$ of input text prompt. Finally, we UV parametrize the garment mesh and project texture information from $\pi_{rgb}$ onto the UV texture map to obtain high-quality textured garment mesh.
%
%
%
\begin{figure*}[ht!]
    \centering
    \includegraphics[width=0.7\linewidth]{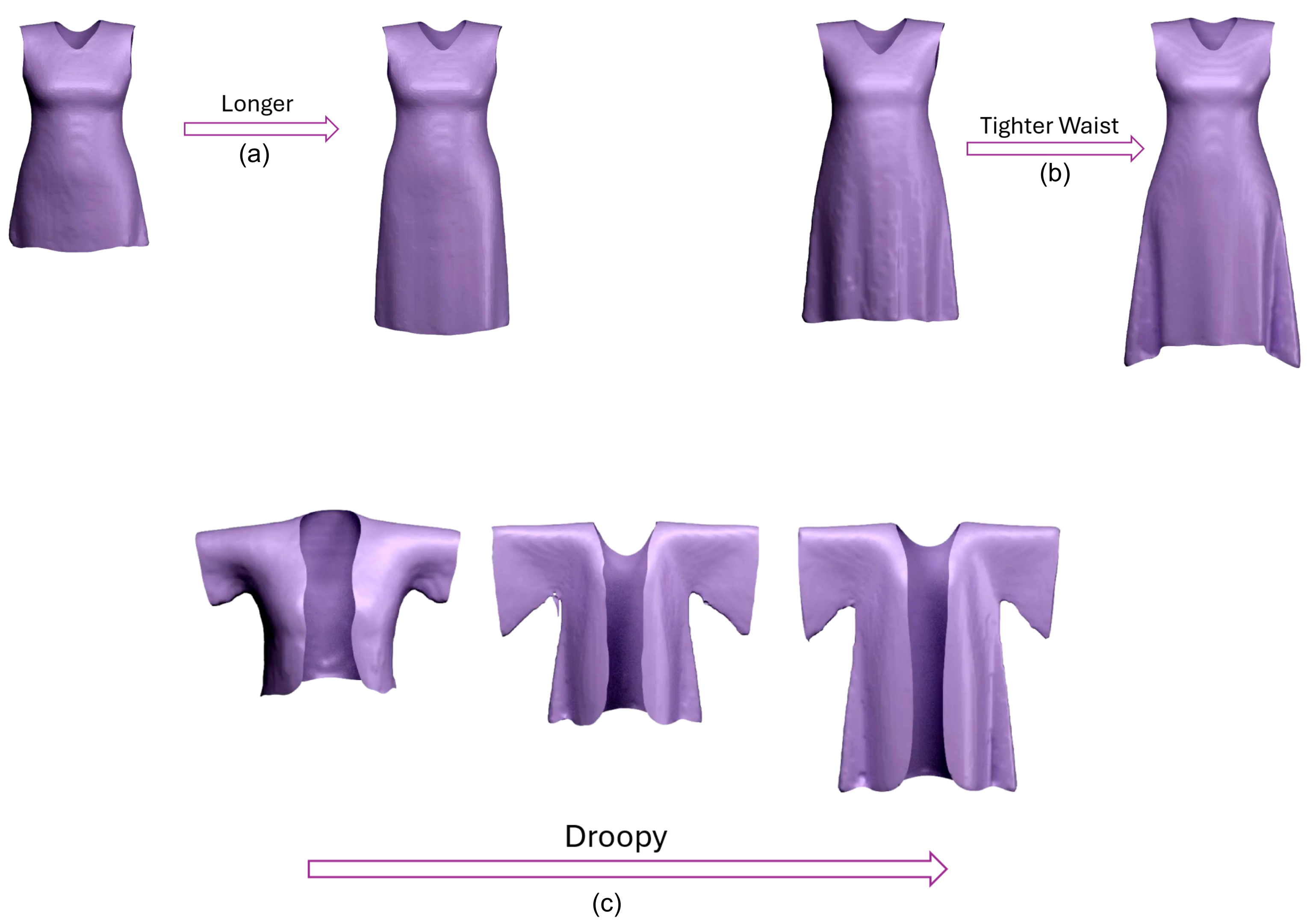}
    \caption{Text-driven manipulation of the latent code.}
    \label{fig:latent_edit}
\end{figure*}
\subsection{Garment Editing via Latent Manipulation}
%
\textit{WordRobe's} encoding enables editing generated garment's attributes by manipulating its corresponding latent code $\phi$. The learned garment space $\Omega$ is continuous and allows meaningful interpolation between different garment latent codes.  As illustrated in \autoref{fig:interp_multiple}, a meaningful garment can be obtained by taking a weighted average of two garment latent codes $\phi_{1}$ \& $\phi_{2}$.

%
%
%
%
%
For \textbf{\textit{text-guided editing}}, we introduce an intuitive approach that uses CLIP arithmetic\cite{patashnik2021styleclip} to automatically identify which dimensions of latent code to manipulate in order to achieve the desired change.
Given a garment latent code $\phi$, methods like \cite{de2023drapenet} employ a pretrained latent code classifier to identify the garment type and to identify which dimension to interpolate in order to induce category-specific changes, e.g. manipulate sleeve lengths. Though this approach works well, it requires explicit manual annotations to enable control over interpolation. In our case, we make use of CLIP embeddings to automatically identify the dimensions to control certain aspects of the 3D garment. Given a CLIP embedding $\psi$ corresponding to a text-prompt (say, \textit{``skirt''}), $MLP_Map$ predicts corresponding garment latent code $\phi$ which is decoded to obtain the initial garment. For a text prompt \textit{``longer''}, we first compute the CLIP embedding $\psi_{feature}$. Then, we perform the following operation: 
\begin{equation}
    \psi_{edit} = w\psi + (1-w)\psi_{feature}
\end{equation}
\noindent
In other words, we perturb the $\psi$ in the direction of $\psi_{feature}$ ($w=0.5$), which results in the modified CLIP embedding $\psi_{edit}$. $MLP_{map}$ takes this $\psi_{edit}$ and predicts $\phi_{mod}$, which if decoded results in the 3D garment which resembles \textit{``long skirt''}. However, predicting the entire garment's latent code from scratch disturbs other characteristics of the original garment as well. To retain the other characteristics as much as possible and change only the \textit{length}, we first compute the difference $\phi_{delta}$ between original latent code $\phi$ and modified latent code $\phi$, i.e. $\phi_{delta}=\phi - \phi_{mod}$. We then identify top-$k$ values in $\phi_{delta}$, which denote the corresponding dimensions to change in order to achieve the desired modification. Thus, perturbing the garment latent code $\phi$ along only these $k$ dimensions results in intended modification while maintaining other characteristics of the garment (choosing the value of k is flexible and is driven by the user preference, we use $k=7$). In \autoref{fig:latent_edit}, we perform text-driven editing by increasing the length of a dress using the prompt \textit{``longer''}, reducing the waist-size of another dress using the prompt \textit{``tighter waist''} and introduce different scale of \textit{droopiness} in the hoodie.
The aforementioned approach does not require any manual intervention except for the amount of perturbation the end-user intends to introduce.


%% file: sec/5_evaluation.tex
\section{Experiments \& Results}
\begin{figure}[h!]
    \centering
    \includegraphics[width=\linewidth]{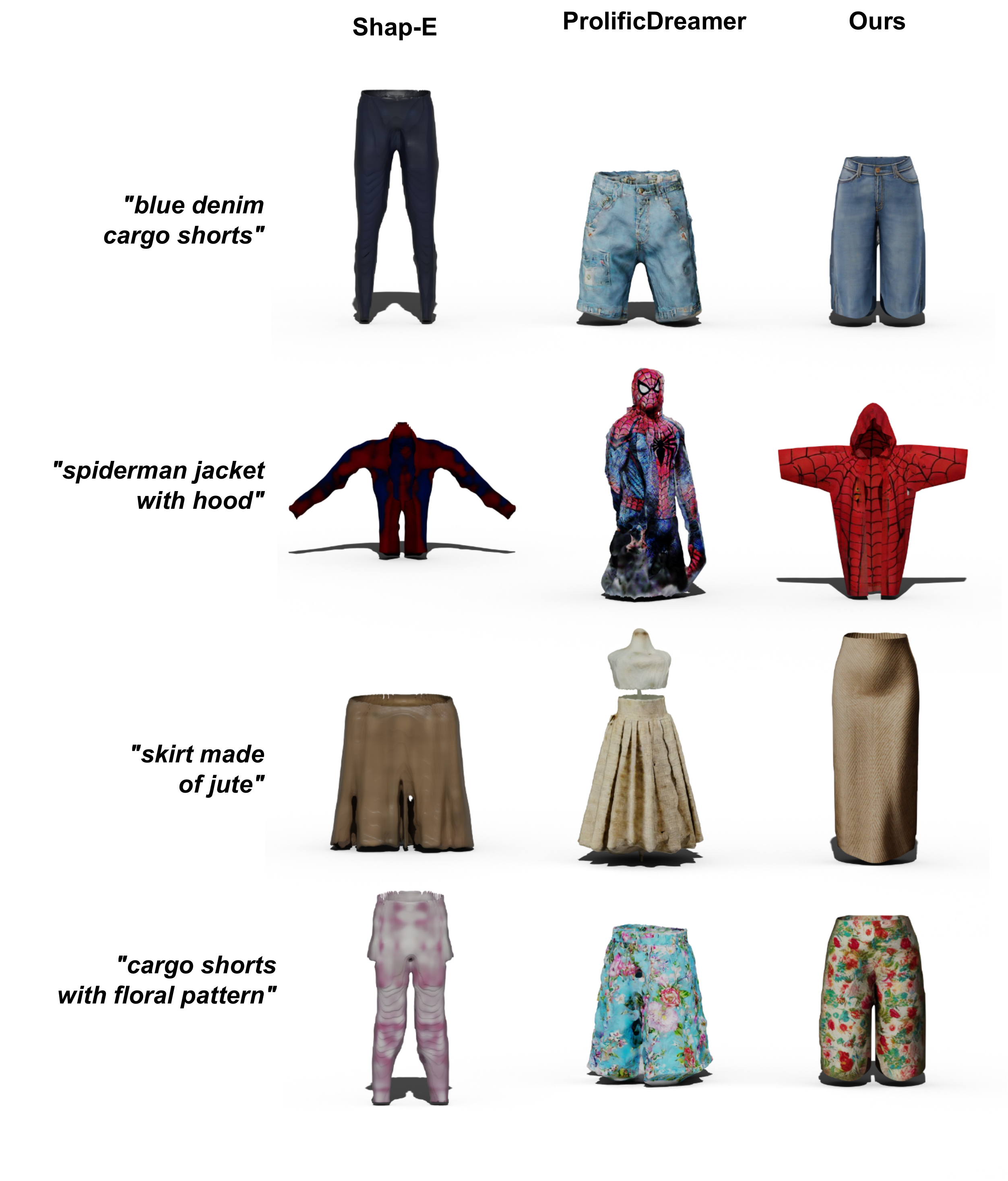}
    \caption{Qualitative comparison with exiting SOTA text-to-3D methods.}
    \label{fig:text_to_3d_SOTA}
\end{figure}

We design several experiments and perform thorough qualitative \& quantitative evaluations of our method, including ablative analysis. Regarding comparison with SOTA methods, since there is no existing method for direct text-driven \textit{unposed} \& \textit{textured} garment generation, we individually compare our two-stage encoder-decoder framework with DrapeNet~\cite{de2023drapenet} and text-guided texture synthesis method with Text2Tex~\cite{chen2023text2tex}, both being SOTA in their respective tasks. We also qualitatively compare with existin SOTA text-to-3D methods in \autoref{fig:text_to_3d_SOTA}.

 All the experiments are done on the 3D garment dataset proposed in \cite{KorostelevaGarmentData}, which has around $20,000$ \textit{unposed} (canonicalized) garments spanning over $19$ categories. We train our garment encoder-decoder networks on $12$ categories and perform evaluations on the remaining unseen $7$ categories, following the official train-test split provided by the authors of \cite{KorostelevaGarmentData}. We are the first one to demonstrate generalization in learning 3D garment latent space on such a relatively large dataset, about \textbf{$\mathbf{30}$ times larger} than the datasets used by the current SOTA (DrapeNet authors show learning only on $600$ training samples across $7$ categories from  CLOTH3D\cite{bertiche2020cloth3d}).
\begin{figure*}[h!]
    \centering
    \includegraphics[width=\linewidth]{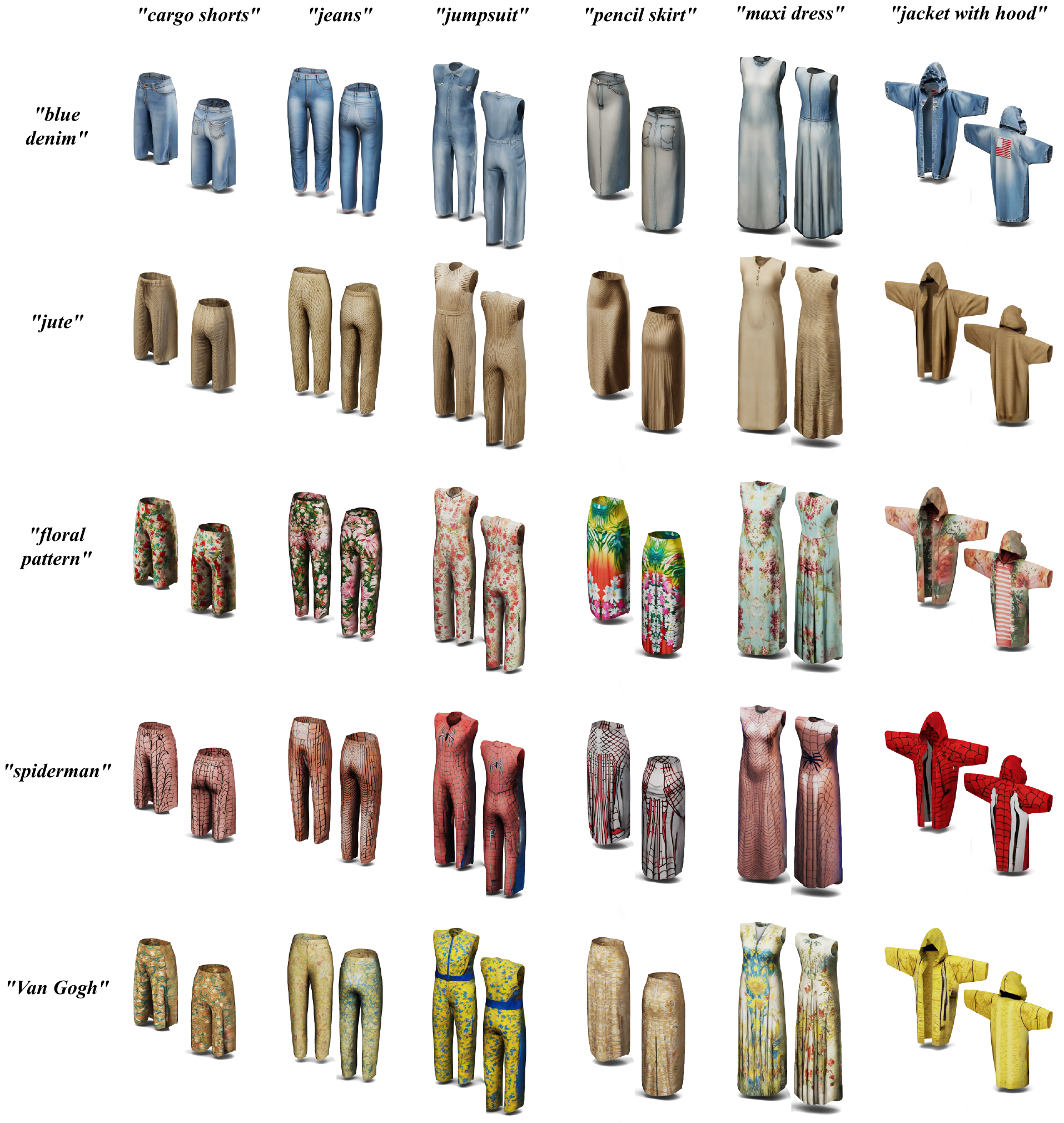}
    \caption{High-quality \textit{unposed} 3D garment meshes with diverse textures generated via \textit{user-friendly} text prompts using \textit{WordRobe}.}
    \label{fig:ours_qual}
\end{figure*}
\subsection{Qualitative Results for Text-Driven Garment Generation}
We visualize 3D textured garment meshes generated using \textit{WordRobe} in \autoref{fig:ours_qual}. As shown in the figure, \textit{WordRobe} generalizes to a large variety of garment styles and textual appearance using \textit{\textit{user-friendly}} text prompts (e.g. \textit{``blue denim cargo shorts''}, \textit{``spiderman jacket with hood'' etc.}). The text prompts in the figure are shortened for readability.
\subsection{Results on Complex Text Prompts}
\begin{figure}
    \centering
    \includegraphics[width=\linewidth]{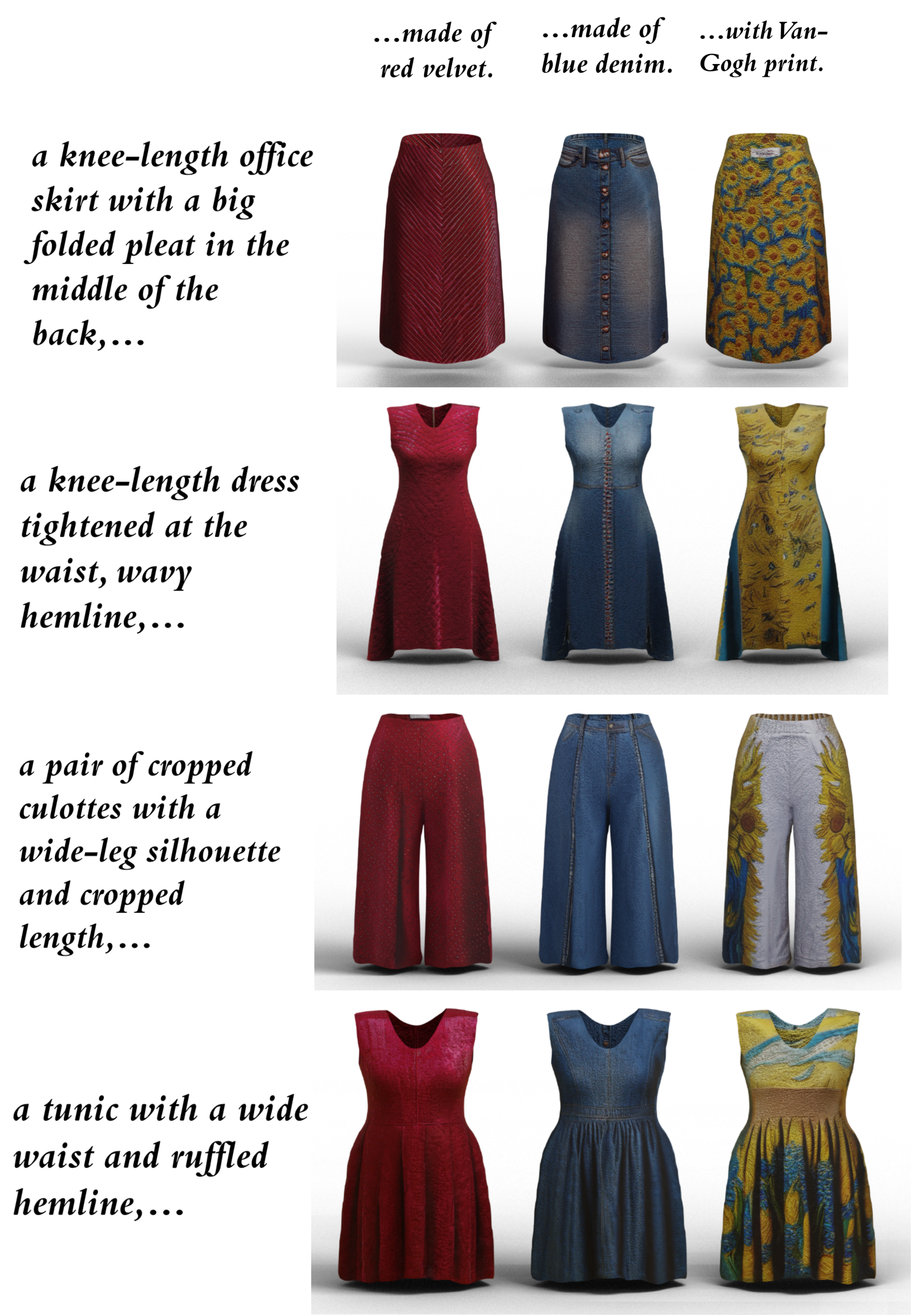}
    \caption{Generating high-quality textured 3D garments with detailed text-prompts.}
    \label{fig:detailed_prompts}
\end{figure}

\begin{figure}
    \centering
    \includegraphics[width=\linewidth]{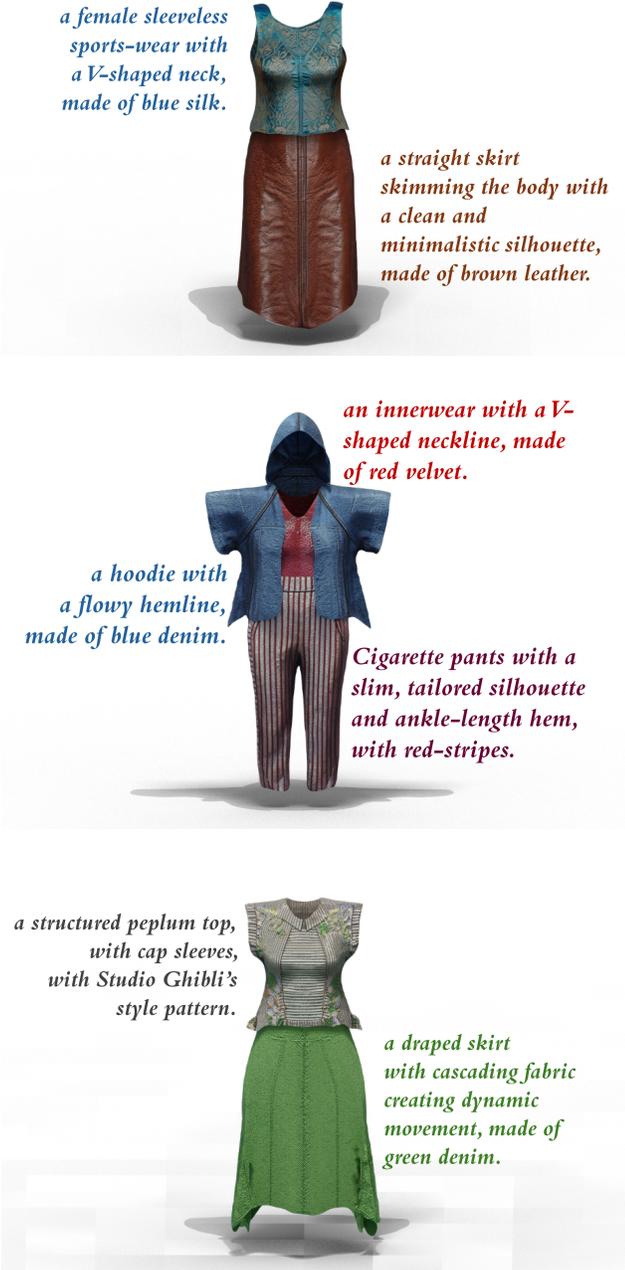}
    \caption{Generating composition of garments using detailed text-prompts.}
    \label{fig:composition}
\end{figure}

Though we aim to generate 3D garments with simplistic, user-friendly prompts, our framework supports detailed and complex prompts as well, as shown in \autoref{fig:detailed_prompts} \& \autoref{fig:composition}. In \autoref{fig:detailed_prompts}, we comprehensively describe the geometry and appearance of the garment to be modelled. Our framework is able to understand detailed prompts and decode them correctly to generate expected geometry and appearance. Similarly, in \autoref{fig:composition}, detailed prompts for separate garments can be given to generate the final combined clothing (by merging the UDF of each garment using \cite{santesteban2021ulnefs}). \textbf{\textit{We recommend watching the supplementary video for high-quality renderings of the results and more such examples of text-driven generation and editing using detailed prompts.}}

%
%
%
\subsection{Sketch-guided 3D Garment Generation \& Editing}
\begin{figure*}
    \centering
    \includegraphics[width=\linewidth]{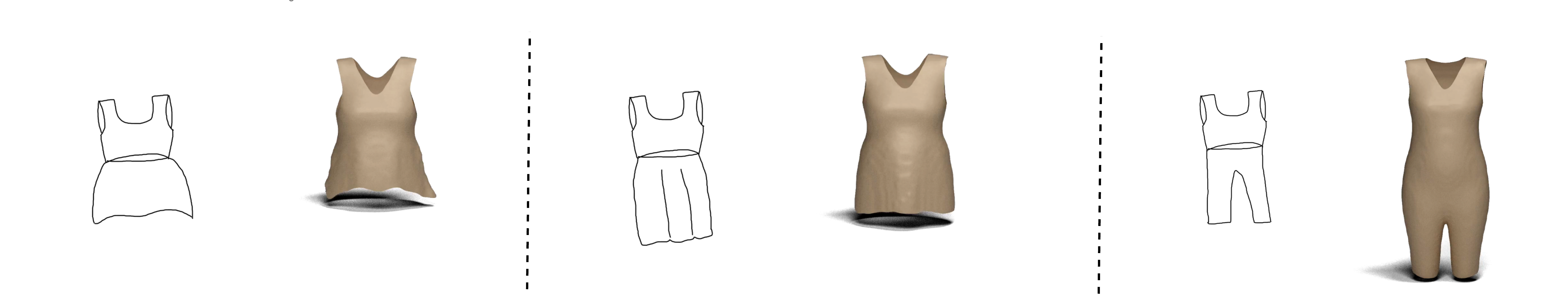}
    \caption{Sketch-guided editing of the 3D garments.}
    \label{fig:sketch_edit}
\end{figure*}

\begin{figure*}
    \centering
    \includegraphics[width=\linewidth]{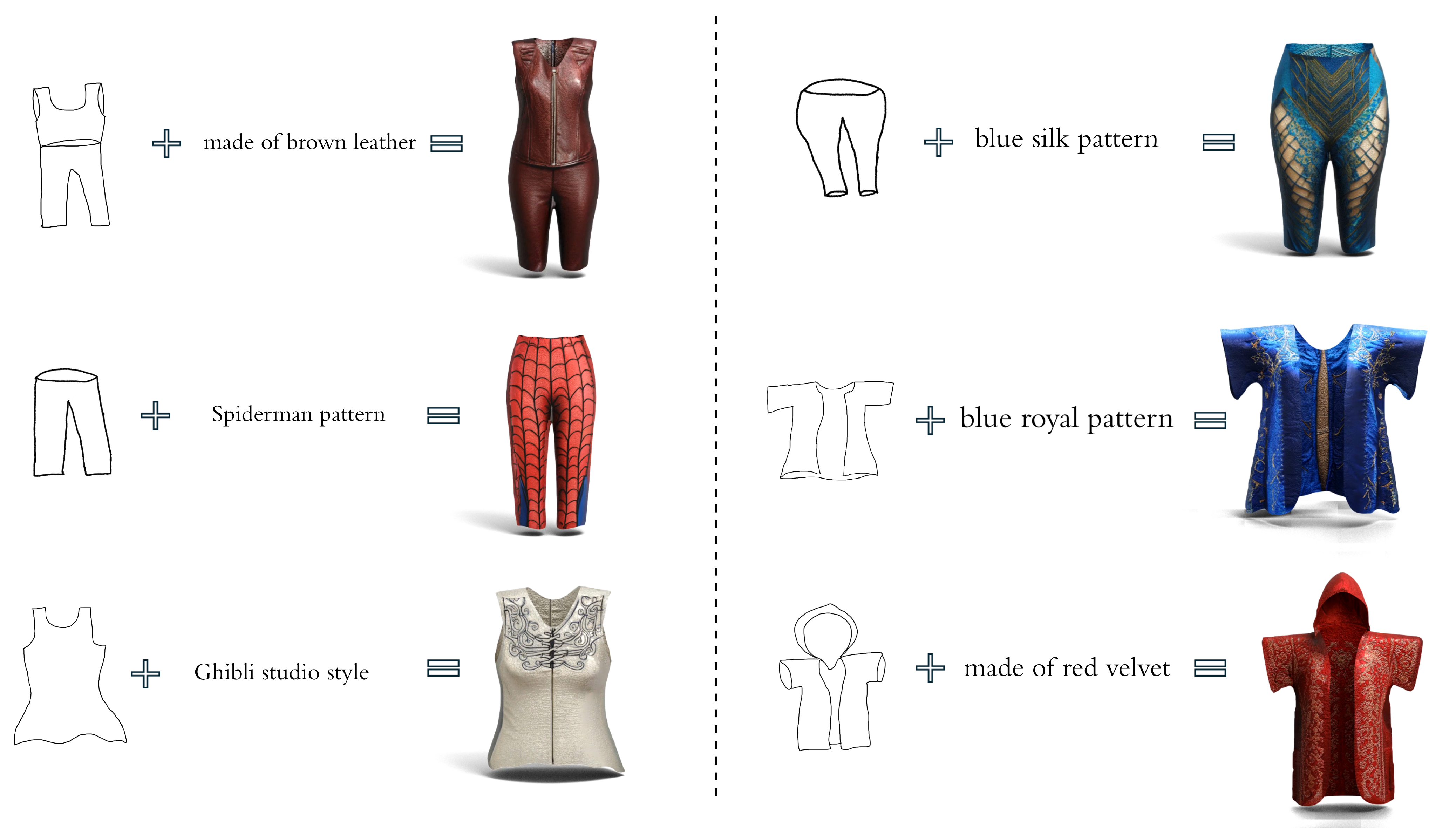}
    \caption{Generating high-quality 3D garments by combining sketch (for geometry) and text (for texture).}
    \label{fig:sketch_gen}
\end{figure*}

Our weakly supervised strategy to train the mapping network on CLIP-embeddings allows us to enable various other interesting applications apart from text-driven generation and editing. Since CLIP space is joint image \& text embedding space, we can pass any image which represents a garment and generate the corresponding 3D garment geometry. \autoref{fig:sketch_edit} demonstrate the way to generate and edit the 3D garments by sketching or scribbling the garments. We pass the garment sketch image to the CLIP and get the corresponding CLIP embedding vector, which is then further passed to the Mapping Network to predict the associated garment latent code. This latent code is then decoded by the coarse and fine decoders to generate the 3D garment geometry. On modifying a part of the sketch, only the corresponding 3D garment part undergoes significant change, while we observe small insignificant changes on the remaining parts of the garments. For textures, we follow the proposed Texture Synthesis module on the 3D geometry obtained via sketches, as shown in \autoref{fig:sketch_gen}. As can be seen from the figure, the geometry of the output 3D garments accurately adheres to the input sketch (at least at a coarser level) in terms of shape and semantics. This feature enables a more controllable and descriptive way to generate the 3D garments compared to text prompts.
\subsection{3D Garments Extraction from Images}
\begin{figure*}
    \centering
    \includegraphics[width=\linewidth]{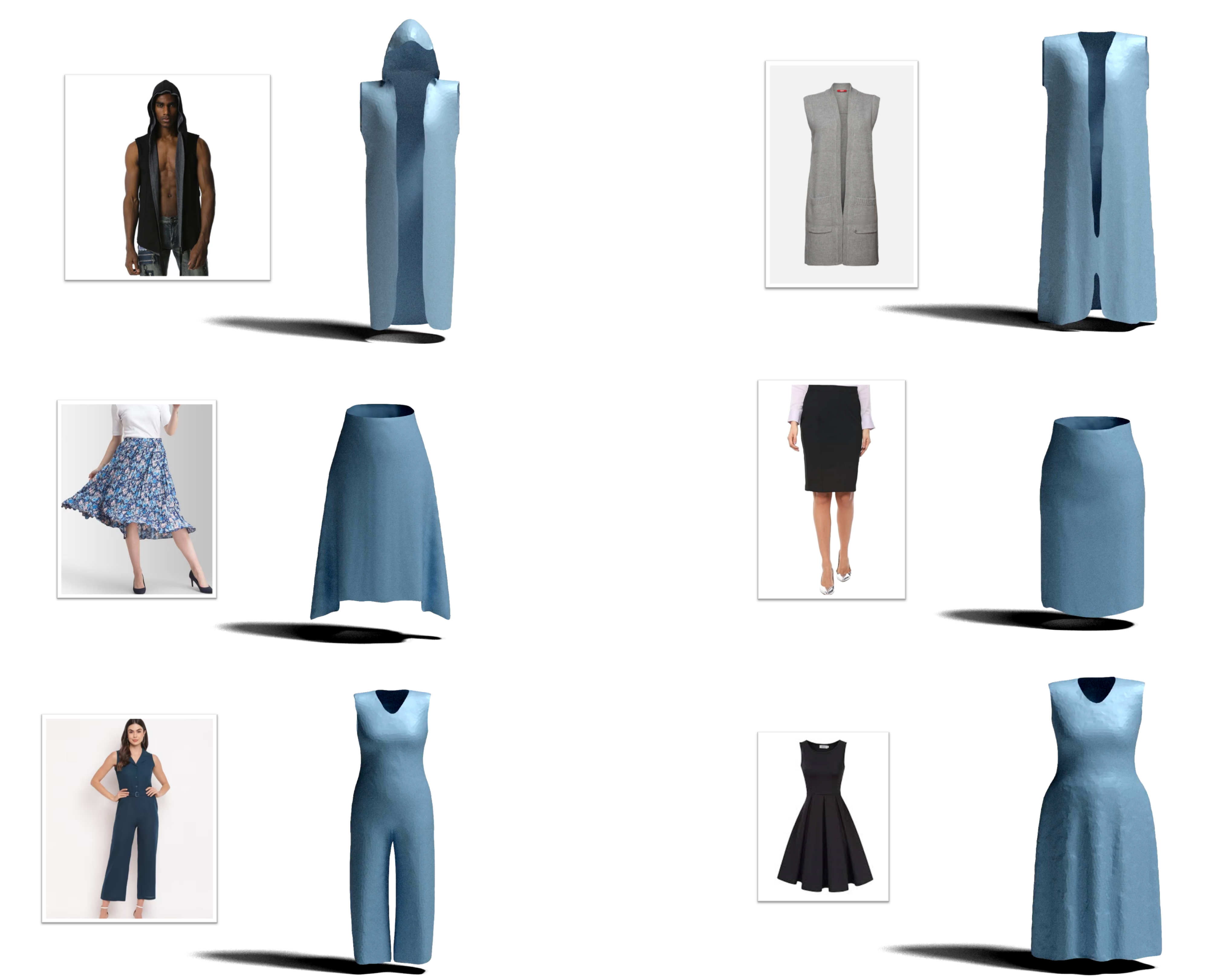}
    \caption{Generating 3D garments using a reference image}
    \label{fig:image_based}
\end{figure*}

We extend the idea from the previous subsection to demonstrate the 3D garment recovery from in-the-wild random garment images. Since we have trained the Mapping Network on data generated by ControlNet, it has seen a wide variety of garment textures and lighting conditions encoded within the CLIP embedding vector. This allows us to pass any internet image to CLIP's image encoder, obtain the corresponding CLIP embedding, and feed it to the Mapping Network to obtain the garment latent code for 3D garment generation. \autoref{fig:image_based} shows the results of the aforementioned approach, where we generate 3D garments from arbitrary garment images whether they contain humans or not.
%
%
\subsection{Evaluation of Garment Latent Space}
\begin{figure}
    \centering
    \includegraphics[width=0.7\linewidth]{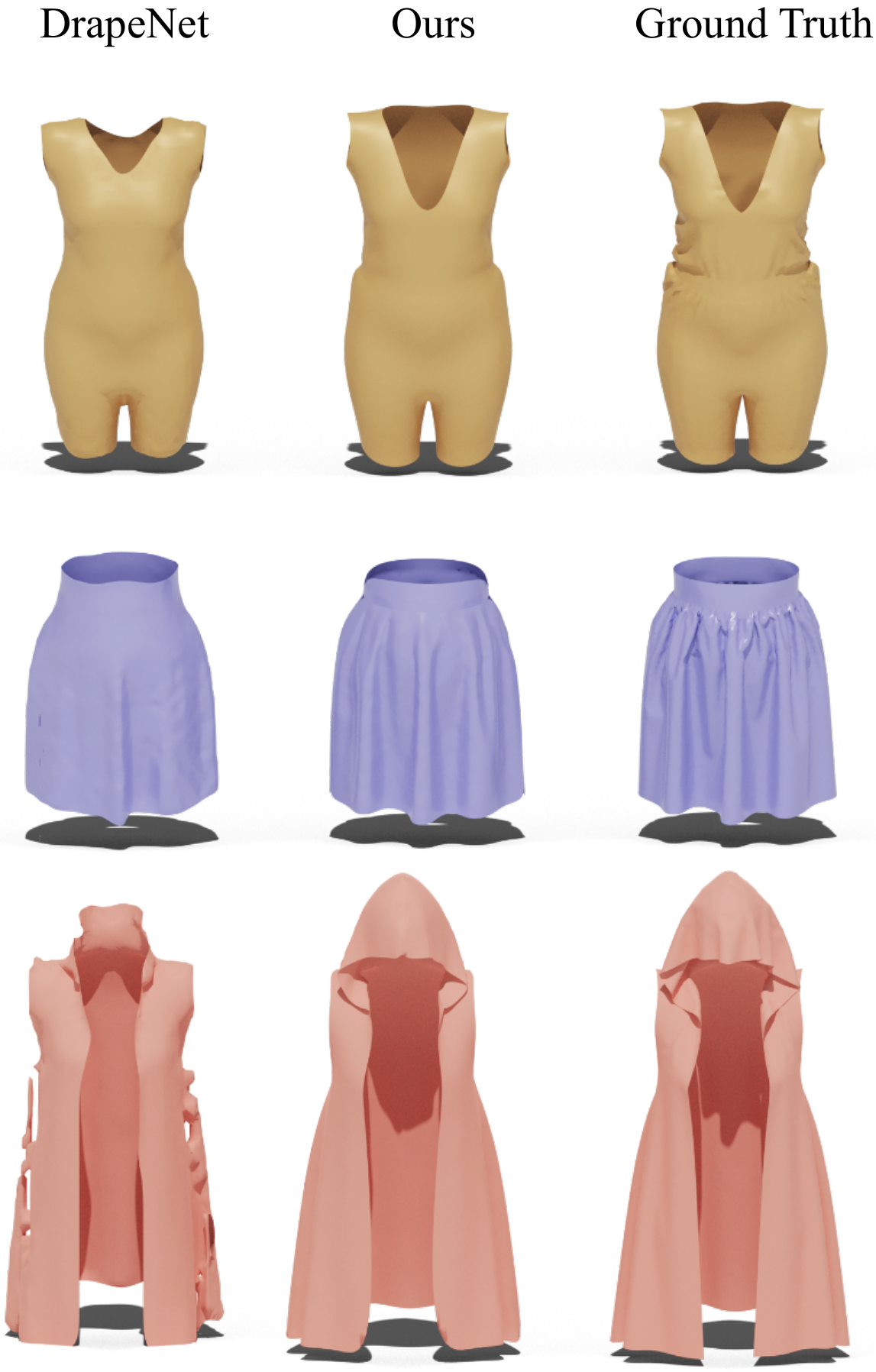}\hspace{2.5em}
    \caption{Qualitative Comparison with DrapeNet~\cite{de2023drapenet}}
    \label{fig:qual_comp}
\end{figure}

\begin{figure}
    \centering
    \includegraphics[width=\linewidth]{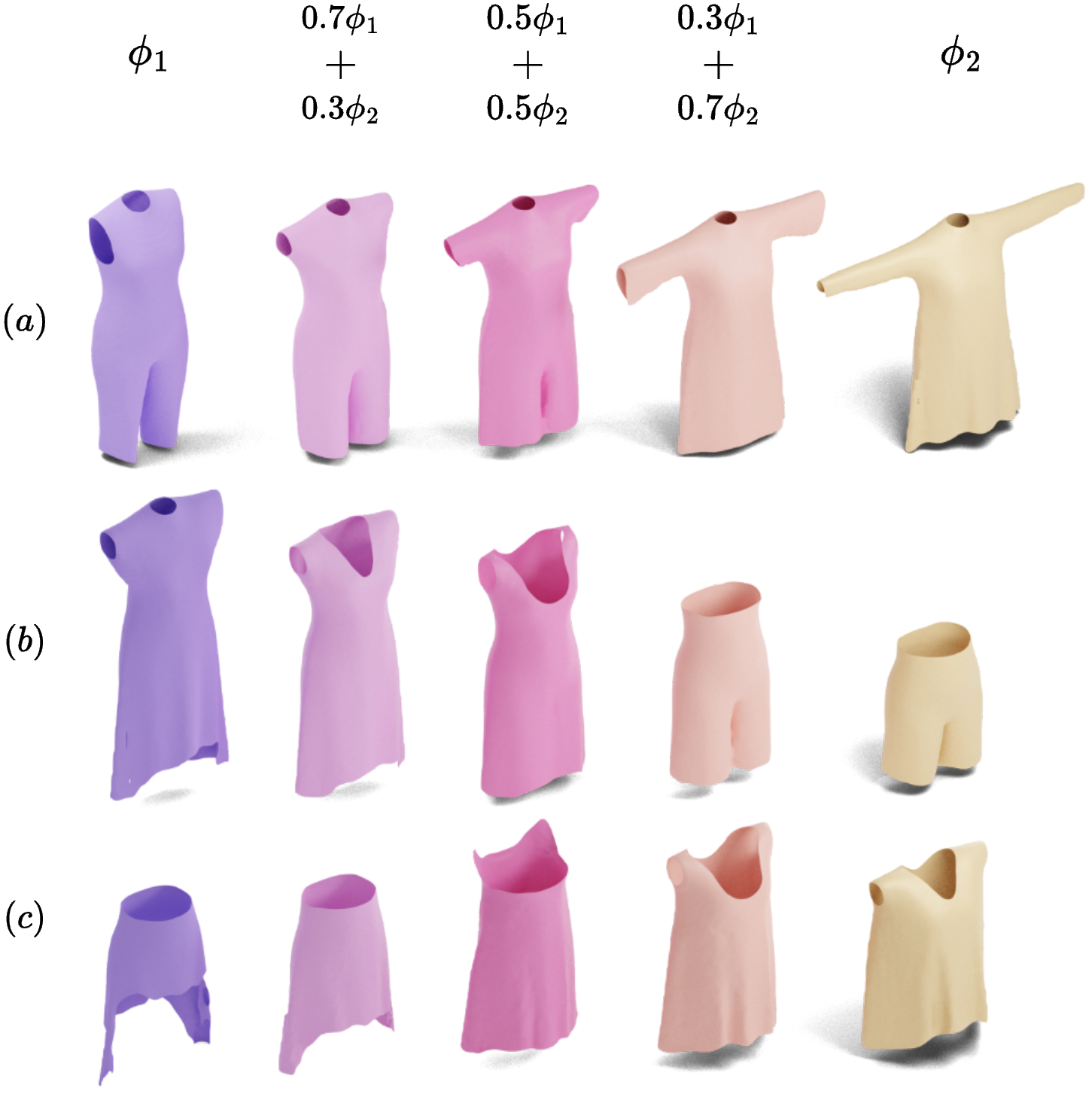}
    \caption{Interpolation of garment latent codes.}
    \label{fig:interp_multiple}
\end{figure}

\noindent
\textbf{{Qualitative Evaluation}:} In \autoref{fig:qual_comp}, we provide a qualitative comparison with DrapeNet\cite{de2023drapenet} on random test samples from unseen garment categories along with the ground truth.
It can be observed from the figure that DrapeNet tends to learn underlying shape, but fails to model garment details. On the other hand, our coarse-to-fine training strategy outperforms DrapeNet in modeling complex garments. We also demonstrate the interpolation capability of the latent space learned using the coarse-to-fine strategy in \autoref{fig:interp_multiple}. For each row $(a)$, $(b)$ \& $(c)$, we first predict two latent codes $\phi_{1}$ \& $\phi_{2}$ using appropriate text prompts and then generate interpolated garments by taking a weighted average of $\phi_{1}$ \& $\phi_{2}$. As shown in the figure, the coarse \& fine decoders decode the interpolated latent code into a meaningful garment geometry.\\
\\
\noindent
\textbf{{Quantitative Evaluation}:} We perform a quantitative comparison of our garment generation method with DrapeNet\cite{de2023drapenet} in \autoref{tab:quant_eval_generation}, where we report standard metrics, Point-to-Surface distance (P2S)\cite{rs11222659} and Chamfer Distance (CD)\cite{bakshi2023nearlinear}, on the test set. We achieve approx. $40$\% lower value for CD and $42$\% lower value for P2S, outperforming DrapeNet by a significant margin. This indicates that the surface quality of the generated 3D garments using \textit{WordRobe} is superior to that of DrapeNet.
\\
\\
\noindent
\textbf{Generalization on CLOTH3D:} We train our encoder-decoder architecture for learning the garment latent space on \textit{unposed} 3D garments from \cite{KorostelevaGarmentData} dataset. However, CLOTH3D\cite{bertiche2020cloth3d} dataset is a widely popular choice for existing state-of-the-art methods like \cite{de2023drapenet}.
Therefore, we demonstrate generalization pn both \textbf{topwear} and \textbf{bottomwear} classes of CLOTH3D dataset after training only on \cite{KorostelevaGarmentData} dataset and computing the standard evaluation metrics, namely Point-to-Surface distance (P2S)\cite{rs11222659} and Chamfer Distance (CD)\cite{bakshi2023nearlinear} on the test set of CLOTH3D. We report both the metrics in \autoref{tab:quant_eval_CLOTH3D}, while comparing with DrapeNet\cite{de2023drapenet}, which has been trained specifically on CLOTH3D. As can be observed from the table, we perform at par, if not better, than DrapeNet, even without training our method on CLOTH3D. This justifies that \textbf{(a) }\cite{KorostelevaGarmentData} has a more diverse and better training distribution than widely popular \cite{bertiche2020cloth3d} dataset, in terms of garment geometry learning; and \textbf{(b)} our encoder-decoder is not overfitted to the samples in \cite{KorostelevaGarmentData} dataset and can generalize to unseen garment types. However, it is important to note that both the datasets, CLOTH3D and \cite{KorostelevaGarmentData} are synthetically generated, as it is very challenging to capture 3D real-world garments in canonical pose.
\begin{table*}
    \centering
    \caption{Quantitative evaluation of garment encoding-decoding framework on both \textbf{topwear} and \textbf{bottomwear} garments from CLOTH3D dataset. Please note that for this experiment, we train our method on \cite{KorostelevaGarmentData} dataset and evaluate on CLOTH3D, while we train and evaluate DrapeNet on CLOTH3D.}
    \begin{tabular}{ccc}
        \toprule
        \textbf{Method} &  \multicolumn{2}{c}{\textbf{Evaluation on CLOTH3D}}\\
        \midrule
        \textbf{} & \textbf{CD(topwear)}$\downarrow$ & \textbf{P2S(topwear)}$\downarrow$ \\
         DrapeNet (trained on CLOTH3D) & $1.522$  & $\mathbf{0.631}$ \\
         Ours (trained on \cite{KorostelevaGarmentData}) & $\mathbf{1.491}$  & $0.635$ \\
         \midrule
        \textbf{} & \textbf{CD(bottomwear)}$\downarrow$ &\textbf{P2S(bottomwear)}$\downarrow$ \\
         
         DrapeNet (trained on CLOTH3D) & $1.585$  & $0.739$ \\
         Ours (trained on \cite{KorostelevaGarmentData}) & $\mathbf{1.568}$  & $\mathbf{0.703}$ \\
        \bottomrule
    \end{tabular}%
    \label{tab:quant_eval_CLOTH3D}
\end{table*}
\begin{figure}
    \centering
0    \includegraphics[width=\linewidth]{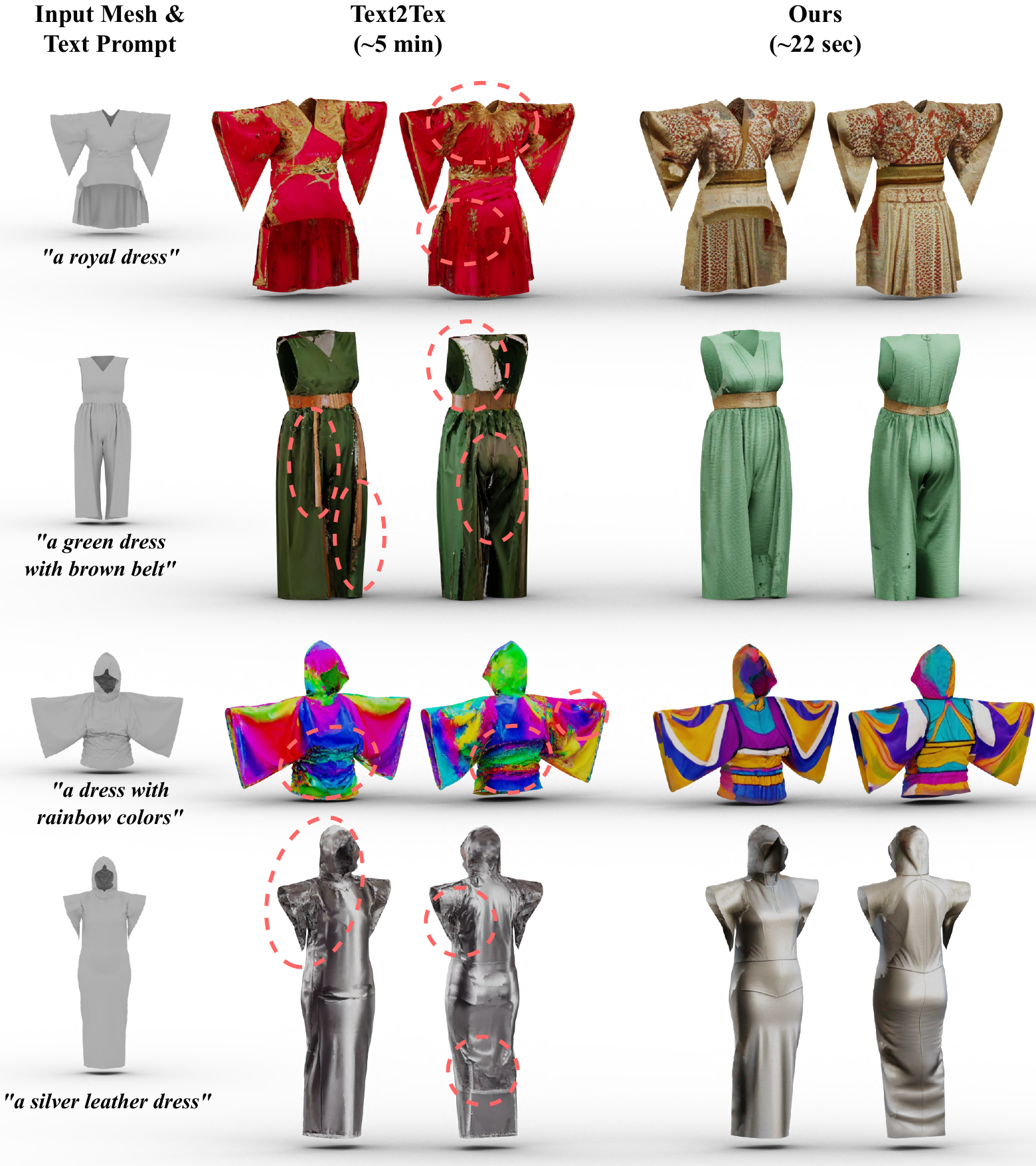}
     \caption{Qualitative comparison of Texture Synthesis. Our method provides better view consistency as compared to Text2Tex (red dotted circle) while being 13 times faster.}
    \label{fig:ours_qual_tex2text}
\end{figure}
\subsection{Evaluation of Texture Synthesis}
\noindent
\textbf{{Qualitative Evaluation}:}
We perform texture synthesis on meshes taken from the test set, governed by the input text prompt from the user and compare with Text2Tex\cite{chen2023text2tex}, as shown in  \autoref{fig:ours_qual_tex2text}. Text2Tex\cite{chen2023text2tex} uses a multiview-optimization strategy, which is slow (takes around $5$ min for a prompt on a single RTX $4090$ GPU) and sometimes converges sub-optimally and results in view inconsistency and patchy artifacts (highlighted in dotted red circles). On the other hand, our optimization-free view-composited method takes around 22 seconds under the same settings, while producing high-quality and view-consistent texture details.
\\
\noindent
\textbf{{Quantitative Evaluation}:} For quantitative comparison with Text2Tex\cite{chen2023text2tex} for generating text-driven textures for a given garment mesh, we first use the texture maps obtained from both methods to render the input garment mesh in $4$ random views. We then compute the average CLIP-Score (higher values are preferred) proposed in \cite{aneja2022clipface} between the input text prompt and the rendered images, and report it in \autoref{tab:clip_score}, where we outperform Text2Tex\cite{chen2023text2tex} under three major variants of the CLIP encoder.
%
%
%
%
%
\begin{table}[t!]
    \centering
    \centering
    \caption{Quantitative evaluation of garment encoding-decoding framework.}
    \begin{tabular}{ccc}
        \toprule
        \textbf{Method} & \textbf{CD}$\downarrow$ & \textbf{P2S}$\downarrow$\\
        \midrule
         DrapeNet\cite{de2023drapenet} & $1.796$  & $0.573$ \\
         Ours* (Single Stage) & $1.631$  & $0.494$ \\
         Ours (w/o $\mathcal{L}_{grad}$) & $1.886$  & $0.612$ \\
          Ours (w/o $\mathcal{L}_{latent}$) & $1.094$ & $0.331$  \\
          \textbf{Ours}& $\mathbf{1.078}$ & $\mathbf{0.329}$   \\
        \bottomrule
    \end{tabular}%
    \label{tab:quant_eval_generation}
\end{table}
\begin{table*}
    \hspace{2em}
    \centering
    \caption{Comparison of text-guided texture synthesis.}
    \begin{tabular}{l c  c  c }
    \toprule
    {\textbf{Method}} & \multicolumn{3}{c}{\textbf{CLIP Score $\uparrow$}}\\
    \cmidrule{2-4}
    & {\textbf{ViT-H/14}} & {\textbf{ViT-L/14}} & {\textbf{ViT-B/16}}\\
    \midrule
    {Text2Tex\cite{chen2023text2tex}} & {$0.263$ ± $0.047$} & {$0.243$ ± $0.041$} & {$0.232$ ± $0.036$}\\
    {\textbf{Ours}} & {$\mathbf{0.304}$ ± $\mathbf{0.043}$} & {$\mathbf{0.265}$ ± $\mathbf{0.037}$} & {$\mathbf{0.241}$ ± $\mathbf{0.034}$}\\
    \bottomrule
    \end{tabular}%
    \label{tab:clip_score}
\end{table*}
%
%
%
%
%
%
\subsection{User Study}
\begin{figure*}
    \centering
    \includegraphics[width=\linewidth]{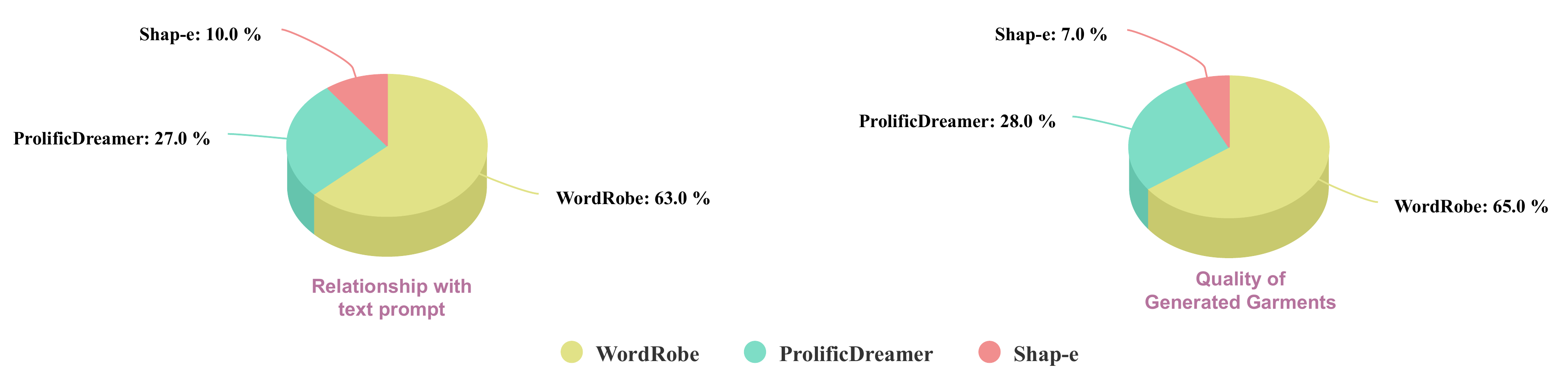}
    \caption{Distribution of preferences in the qualitative user study.}
    \label{fig:user_study_pie}
\end{figure*}
For the subjective evaluation of our method, we perform a qualitative user study among $67$ participants. First, the participants were presented with the results of our method and were asked two questions :
\begin{itemize}
    \item \textbf{How would you rate the relationship between the input text prompt and the generated result on a scale of 1 to 3? [1-not related, 2-somewhat related, 3-highly related]}\\
    \item \textbf{How would you rate the quality of the results in general, on a scale of 1 to 5? [1-very bad, 2-bad, 3-acceptable, 4-good, 5-very good]}
\end{itemize}

Our method achieves an average rating of $\mathbf{2.57}$ on a scale of $[1-3]$ in terms of the relationship between the result and the input text prompt and an average rating of $\mathbf{4.01}$ on a scale of $[1-5]$ in terms of quality of the generated 3D garment.

All the participants were also asked to select one of the methods among \cite{jun2023shape}, \cite{wang2023prolificdreamer} and WordRobe on two basis $-$ relationship between the result \& text-prompt, and overall quality of the generated garment. About $63\%$ of the participants prefer WordRobe, $27\%$ prefer \cite{wang2023prolificdreamer}, and $10\%$ prefer \cite{jun2023shape} in terms of the relationship between the result and text prompt. In terms of quality, $65\%$ of the participants prefer WordRobe, $28\%$ prefer \cite{wang2023prolificdreamer}, and $7\%$ prefer \cite{jun2023shape}. \autoref{fig:user_study_pie} shows the distribution of user preferences in terms of the relationship between the generation \& text prompt, and also in terms of the overall quality of generated 3D garments.

Finally, we also asked participants to choose between Text2Tex\cite{chen2023text2tex} and WordRobe for the text-driven synthesis of textures over existing meshes. About $54\%$ participants opted for WordRobe when it comes to the relationship between the result and input text prompt, and about $76\%$ opted for WordRobe when it comes to the quality of the generated 3D garment.
%
%
%
%
\subsection{Ablation Study}
\begin{figure*}[t!]
    \centering
    \includegraphics[width=\linewidth]{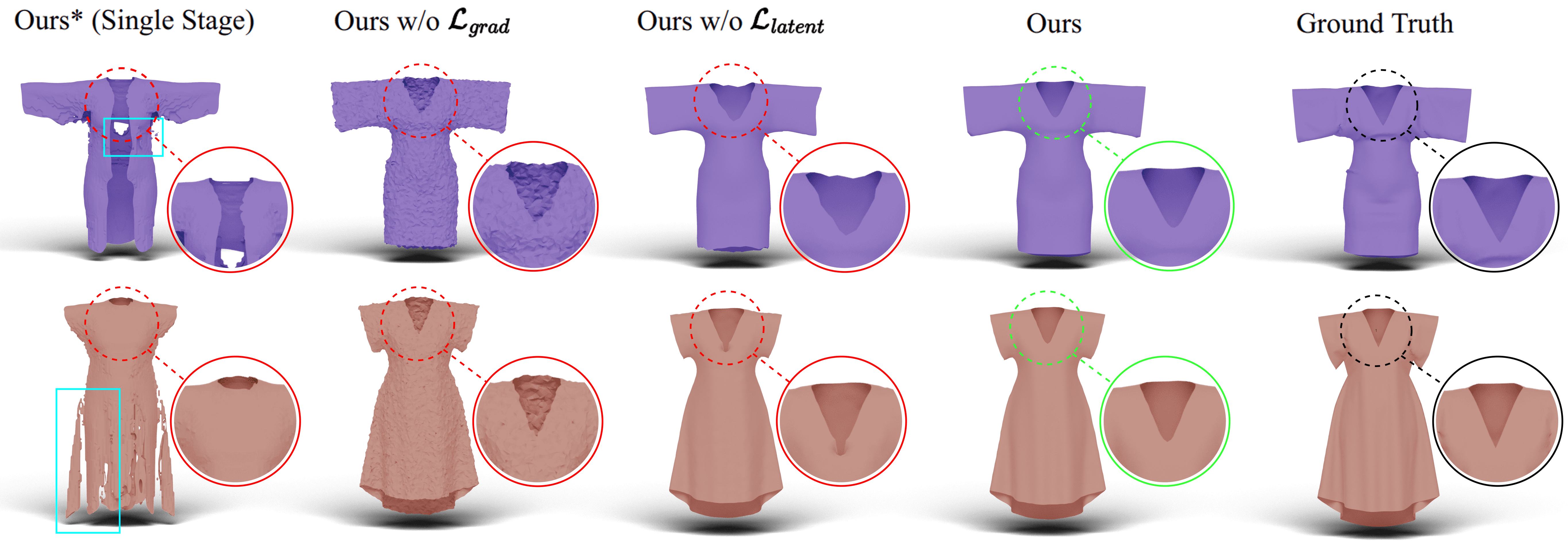}
    \caption{Qualitative ablation of the proposed encoder-decoder framework.}
    \label{fig:qual_ablate}
\end{figure*}
\label{sec:qual_ablate}
\noindent
\textbf{{Single-stage vs Two-stage Decoding}:} We study the importance of two-stage (coarse-to-fine) decoding of the latent code to produce noise-free garment geometry. As shown in \autoref{fig:qual_ablate}, \textbf{Ours*} (proposed framework but with only a single decoder), results in holes \& isolated noise (highlighted in cyan boxes). However, employing both coarse and fine decoders significantly suppresses the noise. This improvement in the surface quality is also evident from \autoref{tab:quant_eval_generation}, where the two-stage framework (Ours) achieves lower values for CD and P2S as compared to single decoder variant \textbf{(Ours*)}.\\
\\
\noindent
\textbf{{Choice of Loss Functions}:} We show a qualitative ablative study on the choice of loss functions used in learning garment latent space in \autoref{fig:qual_ablate} and report the quantitative numbers (CD \& P2S) in  \autoref{tab:quant_eval_generation}. We observe that $\mathcal{L}_{grad}$ plays a significant role in reducing high-frequency surface noise by acting as a regularizer, resulting in lower values of CD and P2S. The use of $\mathcal{L}_{latent}$ provides an improvement in the quality of the garment (especially around the boundaries), however, the drop in CD and P2S values is not very significant. At last, we also perform an ablative study over the choice of losses while training $MLP_{map}$, report Mean Square Error (MSE) over the test set in \autoref{tab:quant_eval_mapping_net}, and conclude that $L1$ loss alone is a more suitable choice for learning a mapping from CLIP space to the garment latent space.\\
\begin{table}
    \centering
    \caption{Quantitative evaluation of training losses for $MLP_{map}$.}
    \begin{tabular}{cc}
    \toprule
     \textbf{Loss} & $\mathbf{MSE_{test}}$ $\downarrow$\\
    \midrule
     $L2$ & $0.7433$\\
    $L1$ + $L_{cosine}$ & $0.3773$ \\
    $L1$ & $\mathbf{0.3481}$ \\
    \bottomrule
    \end{tabular}
    \label{tab:quant_eval_mapping_net}
\end{table}
\begin{table}
    \centering
    \caption{Quantitative evaluation of latent interpolation.}
    \centering
    \begin{tabular}{c c c}
    \toprule
     & $\mathbf{\Delta_{area}}$ $\downarrow$ & $\mathbf{\Delta_{vol}}$ $\downarrow$ \\
    \midrule
     w/o $\mathcal{L}_{latent}$ & $0.028$  & $1.275$ \\
    with $\mathcal{L}_{latent}$ & $\mathbf{0.022}$  & $\mathbf{1.206}$ \\
    \bottomrule
    \end{tabular}
    \label{tab:quant_eval_latent_loss}
\end{table}
\\
\noindent
\textbf{Interpolation Study:} We conduct a quantitative ablation study in \autoref{tab:quant_eval_latent_loss} to understand the effect of $\mathcal{L}_{latent}$ in achieving better interpolation in the garment latent space. Generally, interpolation is assessed qualitatively. Therefore, we formulate two novel metrics for evaluating the interpolation quantitatively, based on the assumption that while interpolating between two shapes, the surface area and volume of the resulting interpolated shape should also get interpolated accordingly \cite{10.1111:cgf.14916}. Given two 3D garments $\mathbb{G}^{1}$ \& $\mathbb{G}^{2}$ and their respective latent codes $\phi_{1}$ \& $\phi_{2}$, the interpolated latent code is obtained as $\phi_{avg}=\alpha\phi_{1}+(1-\alpha)\phi_{2}$, which is then decoded to obtain garment $\mathbb{G}^{avg}$. Here, $\alpha$ is the interpolation weight ranges between $0$ and $1$. We define interpolated surface area difference $\Delta_{area} = || \mathcal{A}(\mathbb{G}^{avg})-\{\alpha \mathcal{A}(\mathbb{G}^{1}) + (1-\alpha)\mathcal{A}(\mathbb{G}^{2})\} ||$ and interpolated volume difference $\Delta_{vol} = || \mathcal{V}(\mathbb{G}^{avg})-\{\alpha \mathcal{V}(\mathbb{G}^{1}) + (1-\alpha)\mathcal{V}(\mathbb{G}^{2})\} ||$, where $\mathcal{A}(\mathbb{G}_{i})$ is the surface area and $\mathcal{V}(\mathbb{G}_{i})$ is the volume (after hole-filling) of the garment mesh $\mathbb{G}_{i}$. We randomly create pairs from test garment meshes and use random values of $\alpha$ for each pair to compute $\Delta_{area}$ \& $\Delta_{vol}$, and report in \autoref{tab:quant_eval_latent_loss}. As evident from the table, the usage of $\mathcal{L}_{latent}$ during training results in lower values of $\Delta_{area}$ \& $\Delta_{vol}$, promoting better interpolation by providing a more organized latent space.
\\
\\
\noindent
\textbf{{Please watch the supplementary video for $\mathbf{360^\circ}$ renderings of the results. }}
\subsection{Training \& Implementation Details}
Our garment generation framework uses DGCNN\cite{dgcnn} as the garment encoder $\xi$ which takes $20,000$ points sampled on the garment surface as input. The decoders $D_{coarse}$ \& $D_{fine}$ are implemented as MLPs, both having 5 hidden layers of 512 neurons each. We also use conditional Batch Normalization \cite{devries2017modulating} for conditioning on the garment latent vector while decoding. We train $\xi$ and $D_{coarse}$ together in the coarse training stage for $20$ epochs, and $D_{fine}$ separately in the fine stage for $10$ epochs. The value of hyperparameters $\lambda_{dist}$, $\lambda_{grad}$ \& $\lambda_{latent}$ involved in training objectives is $1.0$, $0.3$ \& $0.2$, respectively. The mapping network $MLP_{map}$ is modelled using a $10$ layer MLP with a skip-connection from the input layer to $4^{th}$ hidden layer. All the networks are trained using AdamW\cite{loshchilov2019decoupled} optimizer on an NVIDIA RTX $4090$ GPU.

We use OptCuts\cite{Li:2018:OptCuts} for mesh parametrization as it eliminates the possibility of overlap in UV space. For projecting textures, we take a texel of UV map and identify on which face (triangle) of the mesh it lies. Once we have the triangle, we take 3D vertex positions of the triangle and calculate the 3D position of the texel using Barycentric interpolation. This 3D point is then projected on $\pi_{rgb}$ and bilinearly interpolated to get the color information which is then stored into the corresponding texel of the UV map. Doing this process for every texel gives us the final texture UV map and, eventually, texture 3D garment mesh.

%% file: sec/6_conclusion.tex
\section{Discussion} 
\subsection{Why \textit{\textbf{unposed}}?}
\textit{Unposed} simply means garments in canonical T-pose, which is a standard rest pose defined for human characters. We aim towards generating \textit{unposed} 3D garments because it has several advantages. First, the garments are free from any pose-specific deformations, which is undesirable while defining garment characteristics, as symmetries are important in the garment designing process, which gets disturbed when garments undergo pose-specific deformations. Second, standard animation or dynamic character simulation pipelines keep their characters in the canonical pose for rigging and skinning purposes; therefore, it makes sense to have garments also defined in the canonical pose. Additionally, recent learning-based cloth simulation methods \cite{santesteban2022snug, Bertiche_2022, grigorev2022hood} also require garments to be in T-pose/canonical pose, thereby making the 3D garments generated by our framework directly usable in all such scenarios. 

\subsection{Front-back vs Multiview Projection for Texture Synthesis}
\begin{figure}[h!]
    \centering
    \includegraphics[width=\linewidth]{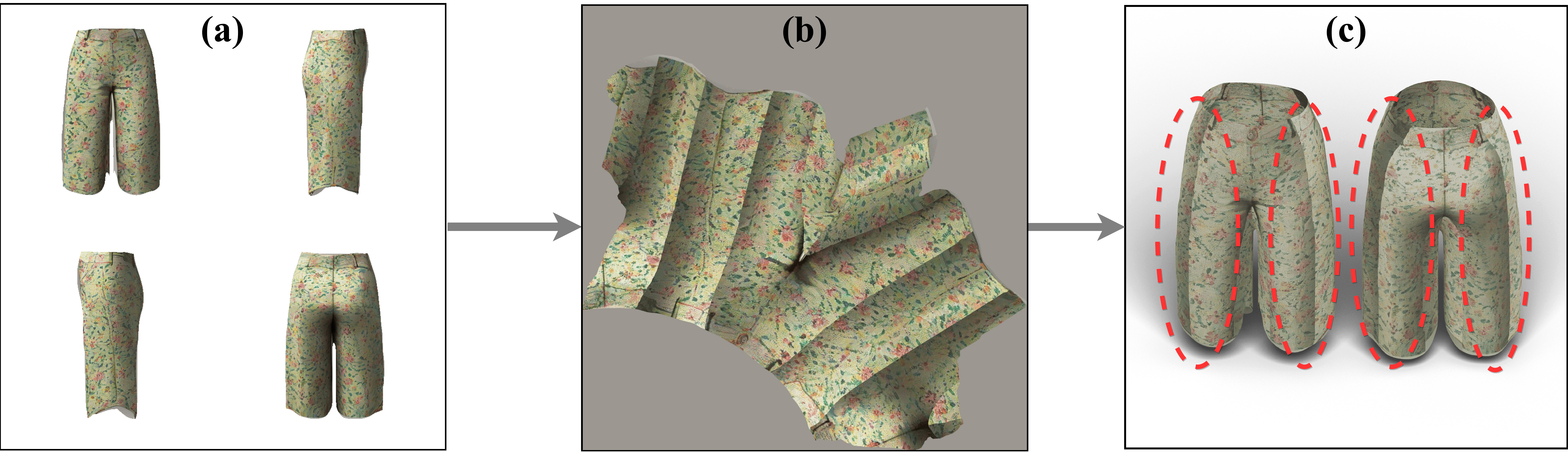}
    \caption{\textbf{(a)} View-composited image (with 4 views) generated using depth-conditioned ControlNet\cite{zhang2023adding}. \textbf{(b)} Corresponding UV texture map with multiple seams. \textbf{(c)} Noticeable seams on the salient parts of the garments.}
    \label{fig:multiview}
\end{figure}
As stated in the main paper, we use front-back as a natural choice for partitioning a 3D garment to reduce the number of visible seams on the mesh. \autoref{fig:multiview} shows a result with $4$ views (instead of just front and back), where prominent seams can be seen (circled in red) on salient regions of the garment mesh. Additionally, we demonstrate better global consistency of the proposed front-back projection via superior CLIP scores (Table.2 main paper), evaluated across \textit{multiple random views}, surpassing SOTA method Text2Tex\cite{chen2023text2tex} (which uses multiview projection), while also being optimization-free \textbf{(\textit{$\mathbf{13}$ times faster})}. This improvement is also justified through qualitative comparison (Figure.8 main paper) and the user study.

\subsection{Contrast with Recent 3D Generative Methods}
\begin{figure}[h!]
    \centering
    \includegraphics[width=\linewidth]{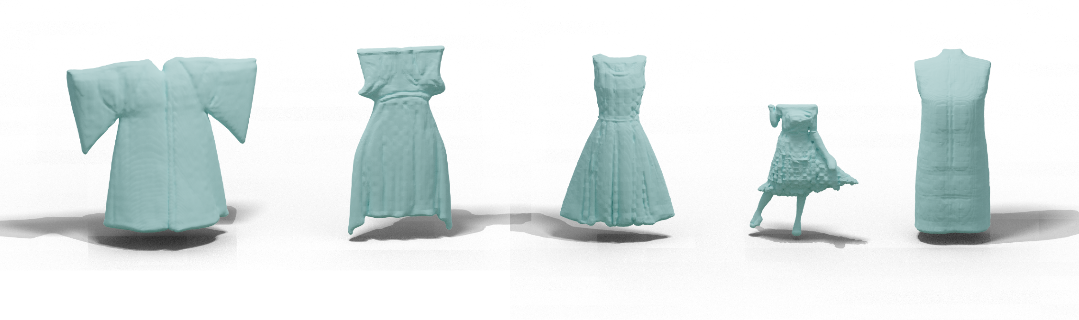}
    \includegraphics[width=\linewidth]{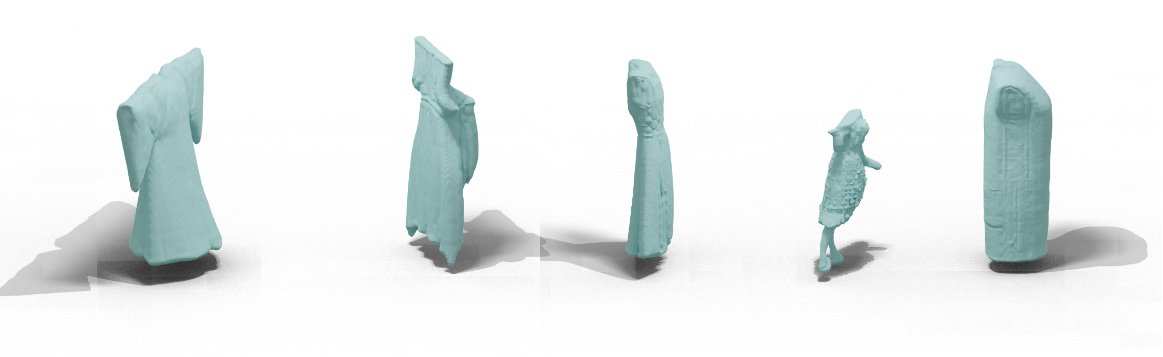}
    \caption{Output from state-of-the-art 3D generative method \cite{tochilkin2024triposr}, after giving just garment images as the input. The geometry from the input view looks plausible (first row) but is poor and unusable when observed from the side (second row).}
    \label{fig:3dgen}
\end{figure}

Recent advancements in 3D generative deep learning have given rise to zero-shot text-to-3D or image-to-3D generative models. However, the geometries obtained from such methods are plausible but nowhere near production-ready, especially when it comes to modelling complex geometries with openings, e.g. garments. Since most of these methods model 3D surfaces as SDFs, they fail to handle open garment surfaces. As shown in \autoref{fig:3dgen}, we highlight the output quality of a recent state-of-the-art 3D generative method \cite{tochilkin2024triposr}, where noisy and poor garment geometries can be seen. Contrast that with the output of our method in \autoref{fig:image_based} on the same garment images, which generate plausible and ready-to-use 3D garments. This performance difference is due to the obvious fact that our method is specialized for 3D garments as compared to generic text-to-3D or image-to-3D methods, which can model arbitrary objects but with poor geometric quality.

\subsection{Text Prompts for Evaluation}
Due to the lack of any 3D garment dataset with text annotations, in order to come up with several diverse text prompts describing the geometry and appearance of the garments, we leverage large language models with powerful language generation capabilities. We asked ChatGPT-3.5 to generate $300$ random text prompts describing different garment styles while just focusing on the geometry and \& $300$ text prompts describing textures only. 
More specifically, after several trials and errors, we came up with the following two prompts:

\begin{itemize}
    \item \textit{``Write $300$ descriptive text prompts to describe various clothing styles. Don't describe color or texture information, just the valid geometrical details, such as size, shape, curvature and so on (do not include knit-type or material type). Remove bullet points and put every point in a new line. Make sure prompts are highly diverse and distinct from each other.''}\\
    \item \textit{``Write $300$ descriptive text prompts to describe various textures and patterns that can be put onto clothing. Be creative and make sure to take inspiration from famous fashion designers. Remove bullet points and put every point in a new line. Make sure prompts are highly diverse and distinct from each other.''}
\end{itemize}

\noindent
We manually verified the correctness, quality and diversity of the text prompts. We report the CLIP score and conduct the user study on the prompts generated using the aforementioned approach. We will release the exact prompts along with the source code in the public domain.


\subsection{Limitations \& Future Directions}
\begin{figure}
    \centering
    \includegraphics[width=0.7\linewidth]{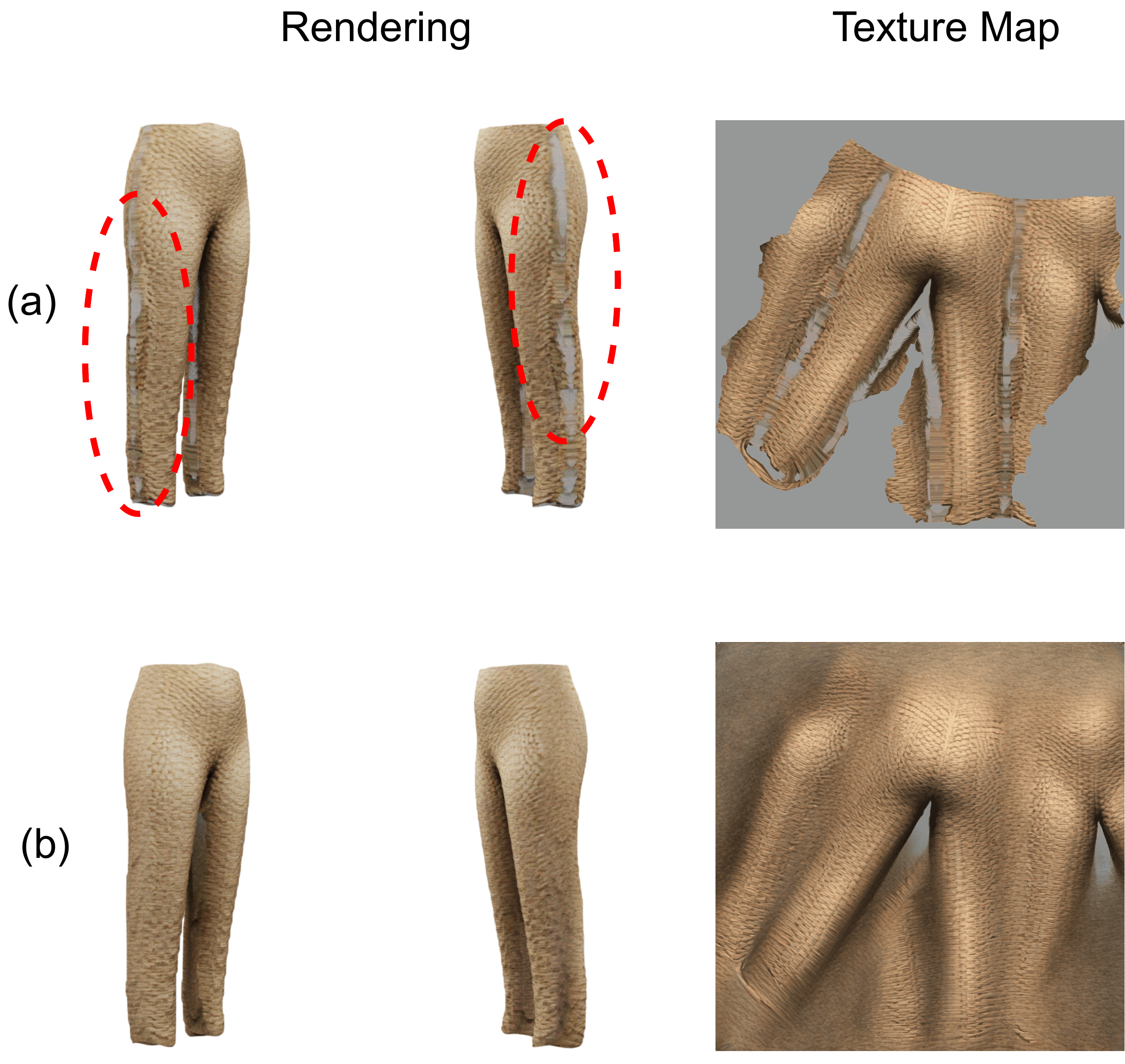}
    \caption{(a) Missing tangential information due to orthographic projection. (b) Details inpainted within the texture map.}
    \label{fig:inpaint}
\end{figure}
WordRobe generates high-fidelity 3D garments with high-quality textures at an unprecedented speed and scale. However, there are certain limitations to our approach that we wish to overcome in future work. One of the drawbacks of front-back orthographic rendering is the loss of information around the tangential regions (see \autoref{fig:inpaint}). In order to fill in the missing details, we employ an off-the-shelf inpainting method, which occasionally leaves blurry seams along the boundary of the front \& back regions. Another limitation is the lack of fine-grain geometrical details on garment parts (e.g. pockets, buttons, etc. ) which makes it challenging to model using implicit representations such as UDF.

In the context of text-driven texture synthesis, one major limitation that every existing method encounters is the hallucination of shadows, lights, and edges, which are purely textural and not a part of the surface of the garment. Though it may enhance the garment's appearance, but from the rendering point of view, this limits the applicability of the extracted textures to new lighting environments. As a part of future work, we wish to explore relighting to get rid of false shading, retaining the true albedo of the garment geometry. We would also like to enable support for layered clothing and material property of the garments.
\section{Conclusion}
We propose \textbf{\textit{WordRobe}}, a novel method for text-driven generation and editing of textured 3D garments. \textit{WordRobe} achieves SOTA performance in learning a 3D garment latent space and in generating view-consistent high-fidelity texture maps. The carefully designed two-stage decoding strategy helps in generating high-quality garment geometry, and the new disentanglement loss promotes better interpolation. Our weakly supervised CLIP-to-latent mapping technique enables text-driven garment generation without requiring any annotated dataset. We report superior qualitative \& quantitative performance compared to existing methods and explain our design choices with appropriate ablative analysis. We believe our work paves the way towards production-ready \textit{unposed} garment generation from text prompts.